\documentclass[letterpaper]{article}
\usepackage[draft]{aaai2026} 
\usepackage{times} 
\usepackage{helvet} 
\usepackage{courier} 
\usepackage[hyphens]{url} 
\usepackage{graphicx}
\usepackage{natbib}
\usepackage{caption}
\usepackage{hyperref}
\usepackage{booktabs}
\usepackage{blindtext}
\usepackage{lipsum}
\usepackage{pifont}
\usepackage{multirow}
\usepackage{subcaption}
\usepackage{amssymb}
\usepackage{relsize}
\usepackage{amsmath}
\usepackage{placeins}

\usepackage{xcolor}
\usepackage{url}

\newcommand{\ours}{InterRVOS\xspace}
\newcommand{\dataset}{InterRVOS-127K\xspace}
\newcommand{\arch}{ReVIOSa\xspace}
\newcommand{\actor}{\texttt{[SEG\_ACT]}\xspace}
\newcommand{\target}{\texttt{[SEG\_TAR]}\xspace}
\newcommand{\cmark}{\ding{51}}
\newcommand{\xmark}{\ding{55}}
\newcommand{\greencmark}{{\color{green!60!black}\ding{51}}}
\newcommand{\redxmark}{{\color{red}\ding{55}}}

\definecolor{qualpink}{HTML}{E07888}
\definecolor{qualblue}{HTML}{6EA8F6}
\newcommand{\qualpink}[1]{{\color{qualpink}#1}}
\newcommand{\qualblue}[1]{{\color{qualblue}#1}}

\usepackage{algorithm}
\usepackage{algorithmic}

\usepackage{newfloat}
\usepackage{listings}
\DeclareCaptionStyle{ruled}{labelfont=normalfont,labelsep=colon,strut=off}
\floatstyle{ruled}
\newfloat{listing}{tb}{lst}{}
\floatname{listing}{Listing}

\title{InterRVOS: Interaction-Aware Referring Video Object Segmentation}

\author {
    Woojeong Jin,
    Seongchan Kim,
    Jaeho Lee,
    Seungryong Kim\footnotemark[2]
}
\affiliations {
    KAIST AI
}

\begin{document}

\maketitle

\def\thefootnote{$\dagger$}
\footnotetext[2]{Corresponding author.}
\def\thefootnote{\arabic{footnote}}

\begin{abstract}
Referring video object segmentation (RVOS) aims to segment objects in a video described by a natural language expression.
However, most existing approaches focus on segmenting only the referred object (typically the actor), even when the expression clearly describes an interaction involving multiple objects with distinct roles. For instance, “A \textit{throwing} B" implies a directional interaction, but standard RVOS segments only the actor (A), neglecting other involved target objects (B).
In this paper, we introduce Interaction-aware Referring Video Object Segmentation (\ours), a novel task that focuses on the modeling of interactions. It requires the model to segment the actor and target objects separately, reflecting their asymmetric roles in an interaction. This task formulation enables fine-grained understanding of object relationships, as many video events are defined by such relationships rather than individual objects.
To support this task, we propose a new evaluation protocol that separately evaluates actor and target segmentation, enabling more accurate assessment of the model's ability to distinguish and segment actor and target roles. We also present \dataset, a large-scale dataset with over 127K automatically annotated expressions, including interaction expressions annotated with distinct masks for actor and target objects. Furthermore, we develop \arch, an MLLM-based architecture that introduces interaction-aware special tokens and leverages an attention mask loss to enhance role-specific segmentation. Extensive experiments show that \arch not only outperforms existing baselines on our proposed \dataset evaluation set, but also achieves strong performance on standard RVOS benchmarks.
Our project page is available at: \href{https://cvlab-kaist.github.io/InterRVOS}{\texttt{https://cvlab-kaist.github.io/InterRVOS}}.

\end{abstract}

\begin{figure}[!t]
    \centering
    \includegraphics[width=\columnwidth]{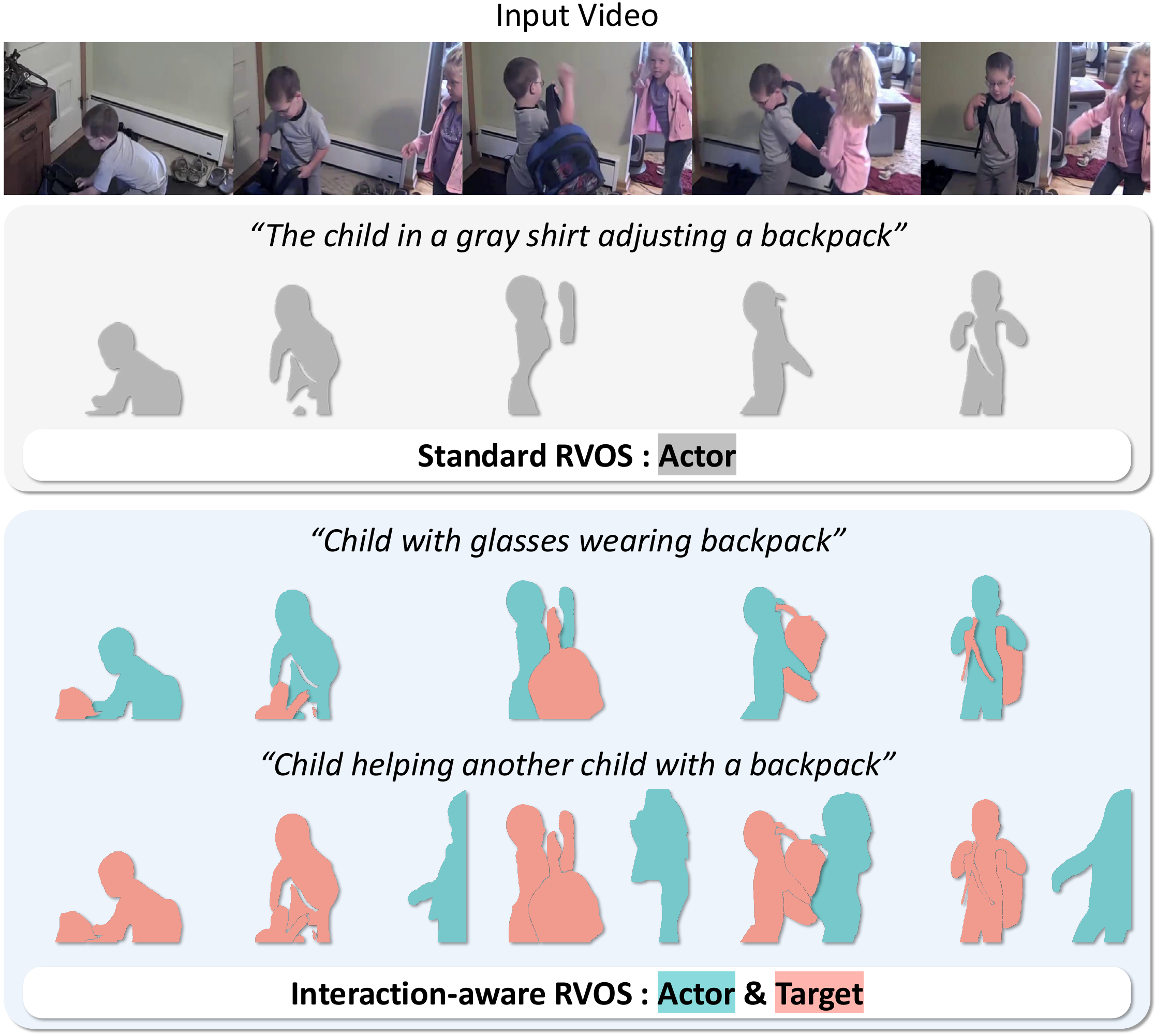}
    \caption{\textbf{Task definition of \ours.}
    We propose a novel task which aims to segment both the actor and the target objects \textit{separately} from a given interaction expression—unlike standard RVOS approaches~\cite{ding2023mevis, wu2022language, wu2022multi, liang2021rethinking, vision-language-transformer, yuan2025sa2va, wang2023soc, zhou2024dshmp, bai2024one} that focus solely on the actor.
    }
    \label{fig:teaser}
\end{figure}

\begin{table*}[t]
    \centering
    \begin{tabular}{l|c|c|cccc}
    \toprule
    Datasets & Annotation & Size & Single & Multiple & Actor-Target \\
    \midrule
    A2D Sentence~\cite{gavrilyuk2018actor} & Manual & 6.6K & \greencmark & \redxmark & \redxmark \\
    J-HMDB Sentence~\cite{gavrilyuk2018actor} & Manual & 0.9K & \greencmark & \redxmark & \redxmark \\
    Ref-DAVIS~\cite{khoreva2019video} & Manual & 1.5K & \greencmark & \redxmark & \redxmark \\
    Ref-Youtube-VOS~\cite{seo2020urvos} & Manual & 15K & \greencmark & \redxmark & \redxmark \\
    MeViS~\cite{ding2023mevis} & Manual & 28K & \greencmark & \greencmark & \redxmark \\
    ReVOS~\cite{yan2024visa} & Manual & 25K & \greencmark & \greencmark & \redxmark \\
    Ref-SAV~\cite{yuan2025sa2va} & Automatic & 72K & \greencmark & \redxmark & \redxmark \\
    \midrule
    \textbf{\dataset} & Automatic & 127K & \greencmark & \greencmark & \greencmark \\
    \bottomrule
    \end{tabular}
    \caption{\textbf{Comparison of existing RVOS datasets and \dataset dataset.} Unlike existing datasets, \dataset additionally supports interaction expressions (denoted as Actor-Target), which annotates separate masks of actor and target objects within an interaction. \dataset is the largest to date (127K mask-text pairs), and is the first to explicitly annotate the masks of actor and targets.
    }
    \label{tab:dataset_comparison}
\end{table*}

\section{Introduction}

Referring Video Object Segmentation (RVOS) aims to segment the object in a video that corresponds to a given referring expression. While earlier works~\cite{seo2020urvos, gavrilyuk2018actor, khoreva2019video, vision-language-transformer, wu2022language, liang2021rethinking, wu2022multi} primarily focused on aligning visual content with language to localize the referred object, recent advancements~\cite{ding2023mevis} have extended the scope of referring expressions to solve more challenging cases, such as motion-only cues or multi-instance references. These trends highlight a growing interest in capturing fine-grained temporal motions and enhancing video-language alignment.

Despite these advances, one important yet underexplored aspect of RVOS is the understanding of \textit{interactions} between objects.
Standard RVOS~\cite{ding2023mevis, wu2022language, wu2022multi, liang2021rethinking, vision-language-transformer, yuan2025sa2va, wang2023soc, zhou2024dshmp, bai2024one} focuses on segmenting a single object or a group of objects exhibiting similar motions, even if expressions that describe interactions with explicit actor and target are given. Such \textit{interaction expressions} include not only the referred objects (actor), but also other objects involved in the interaction (target).
For instance, an expression such as “A \textit{extending a hand towards} B" implies distinct semantic roles and spatio-temporal relationships between objects, where A is the \textit{actor}, and B is the \textit{target}. However, most existing RVOS approaches segment only the actor (A), neglecting the target object (B) involved in the interaction.
Understanding such inter-object dynamics and the ability to distinguish between actor and target roles are essential, as many events and actions in videos are defined not only by the motion of the object itself, but also by its relational context between multiple objects.

In this work, we propose \textbf{Interaction-aware Referring Video Object Segmentation (\ours)}, a novel task that extends standard RVOS by requiring the model to segment the actor and target objects \textit{separately}, as illustrated in Figure~\ref{fig:teaser}.
Importantly, this task explicitly models role directionality within interactions, capturing the asymmetry between actor and target.
This task formulation goes beyond segmenting all involved objects as a whole (i.e., union). It requires the model to separately model each object’s temporal behavior and to capture the inter-object dynamics that arise from their distinct roles.
To enable evaluation under the \ours setting, we introduce a new protocol that assesses segmentation performance separately for actor and target separately, for each interaction expression.

To support this task, we present \textbf{\dataset}, a large-scale dataset containing over 127K expressions, automatically annotated using our data annotation pipeline. Unlike previous RVOS datasets~\cite{seo2020urvos, gavrilyuk2018actor, yan2024visa, khoreva2019video, ding2023mevis}, \dataset includes separate mask annotations for actor and target objects for each interaction expression, enabling models to learn inter-object dynamics effectively.
An overall comparison of datasets is provided in Table~\ref{tab:dataset_comparison}.

We further propose \textbf{\arch}, a novel architecture built upon a multimodal large language model (MLLM). Recent MLLM-based RVOS approaches~\cite{lai2024lisa, yan2024visa, bai2024one, yuan2025sa2va, wang2024villa} utilize \texttt{[SEG]} tokens produced by the MLLM as prompt-like inputs to external segmentation models~\cite{cheng2022masked, ravi2024sam}. Unlike previous methods that use a single \texttt{[SEG]} token, \arch introduces interaction-aware special tokens, \actor and \target, each responsible for segmenting the actor and target objects, respectively.
To further support role-specific segmentation, we introduce attention mask loss (AML), which supervises the attention maps of these tokens to enforce alignment with corresponding object regions.
By guiding the model to attend distinctly to actors and targets, this explicit role separation, enabled by specialized tokens and AML, not only improves the model’s ability to capture inter-object dynamics but also role-specific motion patterns.

To summarize, our main contributions are as follows:
\begin{itemize}
\item We introduce a new task, \ours, which goes beyond the standard RVOS by requiring distinguished segmentation mask of both actor and target objects. We also propose a corresponding evaluation protocol that requires segmenting actor and target objects independently from a single interaction expression.
\item We present \dataset, a large-scale dataset containing over 127K expressions including interaction expressions with distinct actor-target annotations, supporting both interaction-aware and standard referring expressions.
\item We propose \arch, a novel MLLM-based architecture that incorporates interaction-aware special tokens and employs attention mask loss to improve role-specific segmentation required in \ours.
\item \arch achieves state-of-the-art results on \dataset, demonstrating its effectiveness in modeling interactions and a precise understanding of complex temporal motions.
\end{itemize}

\section{Related work}

\begin{figure*}[!t]
    \centering
    \includegraphics[width=\textwidth]{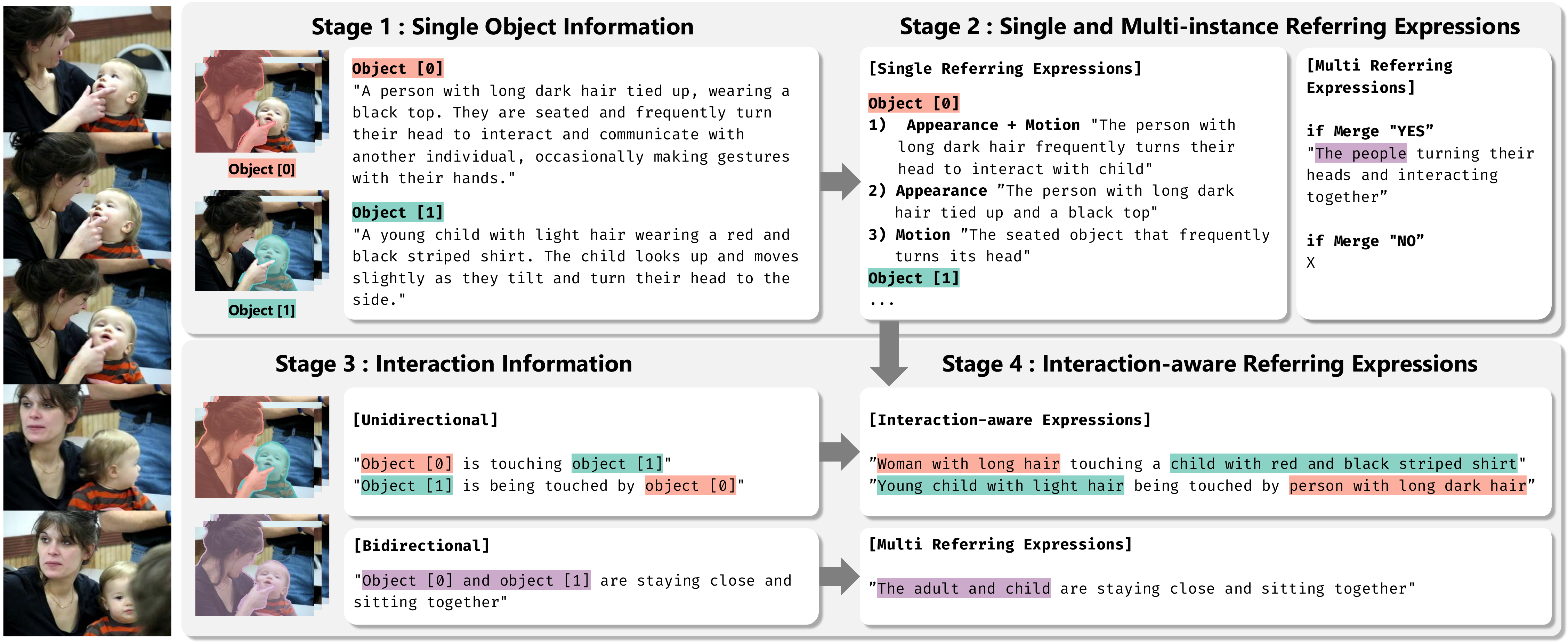}
    \caption{\textbf{Data annotation pipeline.} Our proposed automatic data annotation pipeline constructs referring expressions for single, multi-object, and interaction scenarios in four stages, which extracts object appearance and motion, detects interactions, and generates detailed expressions grounded in both visual properties and interaction context.}
    \label{fig:pipeline}
\end{figure*}

\paragraph{Referring Video Object Segmentation (RVOS).}
RVOS aims to segment a referred object in a video given a natural language expression. Early works~\cite{gavrilyuk2018actor, vision-language-transformer, botach2022end, wu2022language, miao2023spectrum, liang2021rethinking, wu2022multi} mainly focused on appearance-based reasoning through multimodal fusion, often in single-frame or single-object settings. The introduction of MeViS~\cite{ding2023mevis} emphasized the importance of motion-aware and spatio-temporal reasoning by including motion-only and multi-instance expressions, prompting models to better track objects over time. Recent approaches~\cite{zhou2024dshmp, wang2023soc} adopt lightweight text-encoder-based frameworks, while others~\cite{bai2024one, yuan2025sa2va, wang2024villa} leverage multi-modal large language models (MLLMs)~\cite{liu2023visual} and use special tokens (e.g., \texttt{[SEG]}) to guide segmentation.

Despite these advances, existing methods remain actor-focused, performing segmentation solely on an single object (or group of objects) even when interaction expressions involving distinct actor and target roles inherently. While extended tasks like ReasonVOS~\cite{yan2024visa} and Grounded Conversation Generation (GCG)~\cite{munasinghe2025videoglamm, rasheed2024glamm} move beyond traditional RVOS, they fall short in modeling interactions with directions between multiple objects. In particular, GCG treats segmentation as a noun phrase grounding problem without capturing interaction semantics such as role asymmetry.

In contrast, \ours explicitly models the asymmetric roles within interactions by separating actor and target, demanding more precise and role-aware segmentation under interaction-aware expressions.

\paragraph{Video object interaction.}
Modeling object interactions in video requires a role-aware perspective that distinguishes actors from targets, as the semantics of relational events (e.g., \textit{“person pushing cart"}) depend on how one object acts upon another. To support such modeling, prior works have introduced several datasets~\cite{shang2019annotating, shang2017video, ji2020action} with structured annotations, which are actor–predicate–target triplets over time, enabling models to capture visual relationships in dynamic contexts. More recent datasets like STAR~\cite{wu2024star} and MOMA~\cite{fan2021moma} further incorporate temporal grounding and causal structure, capturing complex interactions.

These efforts collectively highlight the importance of explicitly modeling inter-object dynamics as a foundation for fine-grained video understanding.
However, such interaction regarding to actor and target remains overlooked in RVOS, where most approaches treat only-actor setting without considering about target objects involved. \ours addresses this issue by requiring the distinct segmentation of actor and target objects within an interaction.

\begin{figure*}[t]
    \centering
    \includegraphics[width=\textwidth]{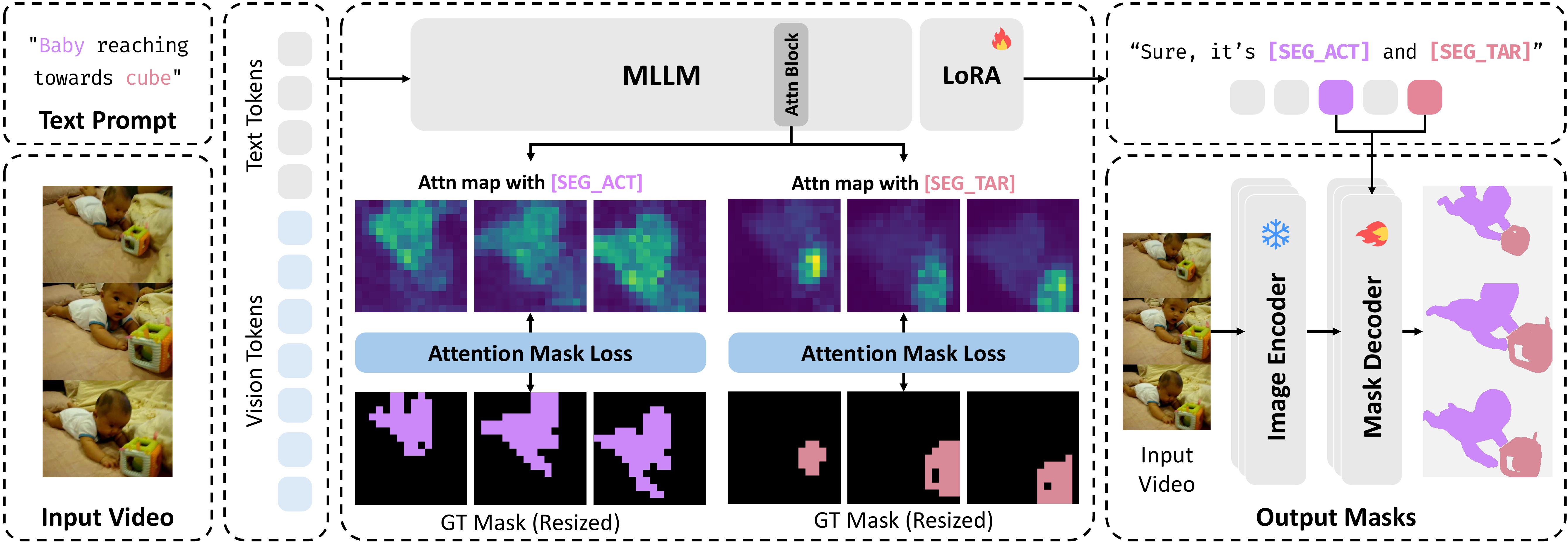}
    \caption{\textbf{Our proposed architecture.}
    Our model utilize \actor and \target tokens which explicitly separate the \textit{actor} and the \textit{target} within an interaction. Furthermore, our model utilize attention mask loss (AML) which enhances the segmentation performance of both the actor and the target, enabling better role separation and ultimately improves interaction modeling.
    }
    \label{fig:architecture}
\end{figure*}

\section{\dataset Dataset}

As \ours requires separate segmentation of actor and target objects, existing datasets~\cite{gavrilyuk2018actor, khoreva2019video, seo2020urvos, ding2023mevis, yan2024visa, yuan2025sa2va} provide limited supervision, particularly lacking in annotations for the \textit{target object}. To address this, we introduce \textbf{\dataset}, an automatically annotated large-scale dataset containing interaction-aware expressions and distinct mask annotations for both actor and target objects.
Built upon VidOR~\cite{thomee2016yfcc100m}, \dataset is constructed via a stage-wise automated annotation pipeline that leverages GPT-4o~\cite{hurst2024gpt} and LLaMA-70B~\cite{grattafiori2024llama} to generate and verify high-quality captions. Additional details on \dataset are provided in the Appendix, including the detailed data annotation pipeline (Appendix~\ref{suppl:annotation_pipeline}), data examples (Appendix~\ref{suppl:data_samples}), the video clip extraction procedure from source videos to the training and evaluation sets (Appendix~\ref{suppl:video_clip}), and overall dataset statistics (Appendix~\ref{suppl:statistics}).

\subsection{Data annotation pipeline}
\label{main:annotation_pipeline}

To generate high-quality expressions which capture the precise interaction between actors and targets, we design a stage-wise automatic annotation pipeline consisting of four main stages. The overall data annotation pipeline is illustrated in Figure~\ref{fig:pipeline}.

Prior to stage-wise processing, we pre-compute mask tracks for all objects in the video using SAM2~\cite{ravi2024sam}. \textbf{Stage 1} captures each object’s appearance and motion independently. \textbf{Stage 2} converts this into referring expressions, optionally merging descriptions for objects with similar motion patterns. \textbf{Stage 3} detects interactions, determines their directionality, and assigns actor and target roles if unidirectional. \textbf{Stage 4} generates rich, interaction-aware expressions by incorporating both class-level and appearance-specific cues, producing multiple paired expressions by swapping actor and target roles. A detailed explanation of the annotation pipeline is provided in the Appendix~\ref{suppl:annotation_pipeline}.

\subsection{Training and evaluation set}

Using data annotation pipeline, we automatically annotated 8,000 videos for training and 738 videos for evaluation. The numbers of expressions are 122,188 and 5,048, respectively. The evaluation set was refined by human annotators, correcting both expressions and segmentation masks.

\section{\arch Architecture}

As \ours emphasizes a detailed understanding of object interactions and diverse motion dynamics, we propose \textbf{\arch} (\underline{\textbf{Re}}ferring \underline{\textbf{V}}ideo \underline{\textbf{I}}nteraction-aware \underline{\textbf{O}}bject \underline{\textbf{S}}egment\underline{\textbf{a}}tion), a novel architecture tailored for this task.
Unlike prior RVOS setting that typically segment only the \textit{actor} object referred to in the expression, \ours requires comprehensive reasoning over the interaction described, explicitly identifying the roles of both the actor and the target, and segmenting them accordingly.
To address these challenges, \arch utilizes interaction-aware special tokens and leverages attention mask loss (AML) to enable accurate disambiguation of actor and target roles and to capture the inter-object dynamics. Furthermore, AML encourages the MLLM to generate segmentation tokens that exhibit stronger aggregation toward the object and enhance vision-language alignment.
The overall architecture of \arch is show in Figure~\ref{fig:architecture}.

\subsection{MLLM-based prompting}

Given an input video $V = \{I_i\}_{i=1}^{T} \in \mathbb{R}^{T \times H \times W \times 3}$ consisting of $T$ frames and a referring expression $E$, our model aims to predict binary segmentation mask sequence $\hat{\mathcal{M}} = \{\hat{\mathcal{M}}_t\}_{t=1}^{T} \in \mathbb{R}^{T \times H \times W}$, where each mask $\hat{\mathcal{M}}_t \in \{0, 1\}^{H \times W}$ corresponds to the objects at time $t$. The overall framework consists of a LLaVA-based~\citep{liu2023visual} multimodal large language model (MLLM) and a video segmentation model, SAM2~\cite{ravi2024sam}.

We first extract vision tokens $\mathbf{f}_{\mathrm{v}} \in \mathbb{R}^{N_v \times D_v}$ from a uniformly sampled video $V'$ consisting of $T'$ frames, and text tokens $\tilde{\mathbf{f}}_{\mathbf{t}} \in \mathbb{R}^{N_t \times D}$ from the referring expression $E$, using the vision encoder and text tokenizer of the MLLM. Here, $N_v$ and $N_t$ denote the number of vision and text tokens, while $D_v$ and $D$ represent the embedding dimensions of the vision encoder and MLLM, respectively. The vision tokens are projected into a shared embedding space with text tokens using MLP projection layer:
\begin{equation}
    \tilde{\mathbf{f}}_{\mathrm{v}} = \mathrm{MLP}_{\mathrm{vision}}(\mathbf{f}_{\mathrm{v}}).
\end{equation}
The projected vision tokens $\tilde{\mathbf{f}}_{\mathrm{v}} \in \mathbb{R}^{N_v \times D}$ and text tokens $\tilde{\mathbf{f}}_{\mathbf{t}}$ are concatenated and fed into the MLLM $\mathcal{F}$ to produce the output sequence $\hat{\mathbf{y}}_{\mathrm{out}}$:
\begin{equation}
\hat{\mathbf{y}}_{\mathrm{out}} = \mathcal{F}([\tilde{\mathbf{f}}_{\mathrm{v}}; \tilde{\mathbf{f}}_{\mathbf{t}}]),
\end{equation}
where $\hat{\mathbf{y}}_{\mathrm{out}}$ includes a special segmentation token, i.e., \texttt{[SEG]}.
We extract the final-layer embedding $\tilde{\mathbf{h}}_{\mathrm{seg}}$ corresponding to the \texttt{[SEG]} token and apply an MLP projection layer, $\mathrm{MLP}_{\mathrm{seg}}$, to obtain the prediction vector $\mathbf{p}_{\mathrm{seg}} \in \mathbb{R}^{D_{\mathrm{dec}}}$, where $D_{\mathrm{dec}}$ is the input embedding dimension of the SAM2 mask decoder.
In parallel, the vision encoder of SAM2 extracts visual features $\mathbf{v}_{\mathrm{seg}} \in \mathbb{R}^{T \times N_\mathrm{enc} \times N_{\mathrm{enc}} \times D_{\mathrm{enc}}}$ from the input video $V$, where $N_{\mathrm{enc}} \times N_{\mathrm{enc}}$ denotes the spatial resolution of the encoder feature map and $D_{\mathrm{enc}}$ is the corresponding feature dimension.

Finally, SAM2 mask decoder $\mathcal{F}_{\mathrm{dec}}$ produces the binary mask sequence $\hat{\mathcal{M}}$. The overall process is formulated as:
\begin{align}
\mathbf{p}{_\mathrm{seg}} = \mathrm{MLP}_{\mathrm{seg}}(\tilde{\mathbf{h}}_{\mathrm{seg}}), \quad
\hat{\mathcal{M}} = \mathcal{F}_{\mathrm{dec}}(\mathbf{v}_{\mathrm{seg}}, \mathbf{p}_{\mathrm{seg}}).
\end{align}

\subsection{Interaction-aware special tokens}
To effectively model inter-object dynamics and enable role-specific segmentation of the actor and target within an interaction, we extend the standard \texttt{[SEG]} token formulation by introducing two interaction-aware special tokens: \actor and \target, representing the \textit{actor} and \textit{target} objects, respectively. By adapting these tokens, the model learns to distinguish between the semantic roles of the involved objects, which implicitly enhances its ability to understand and recognize complex interactions in more precise.

Depending on the type of referring expression $E$, the model dynamically determines whether to generate one or two special tokens.
At inference time, the model first determines whether the input expression involves an interaction. If so, it outputs both the \actor and \target tokens for distinct segmentation. Otherwise, only the \actor token is generated for actor segmentation.
In this new setting, the output of MLLM $\hat{\mathbf{y}}_{\mathrm{out}}$ can now include interaction-aware special tokens.

The corresponding hidden states for each special token $\tilde{\mathbf{h}}_{\mathrm{act}}$ and $\tilde{\mathbf{h}}_{\mathrm{tar}}$ at the last layer of the MLLM are projected into SAM2’s prompt embedding space:
\begin{align}
    \mathbf{p}_{\mathrm{act}} = \mathrm{MLP}_{\mathrm{seg}}(\tilde{\mathbf{h}}_{\mathrm{act}}), \quad
    \mathbf{p}_{\mathrm{tar}} = \mathrm{MLP}_{\mathrm{seg}}(\tilde{\mathbf{h}}_{\mathrm{tar}}),
\end{align}
where $\mathbf{p}_{\mathrm{tar}}$ is used only when \target is generated.
Finally, the segmentation mask outputs are computed as:
\begin{align}
    \hat{\mathcal{M}}_{\mathrm{act}} = \mathcal{F}_{\mathrm{dec}}(\mathbf{v}_{\mathrm{seg}}, \mathbf{p}_{\mathrm{act}}),
    \quad
    \hat{\mathcal{M}}_{\mathrm{tar}} = \mathcal{F}_{\mathrm{dec}}(\mathbf{v}_{\mathrm{seg}}, \mathbf{p}_{\mathrm{tar}}).
\end{align}

\begin{figure}[!t]
    \centering
    \includegraphics[width=\columnwidth]{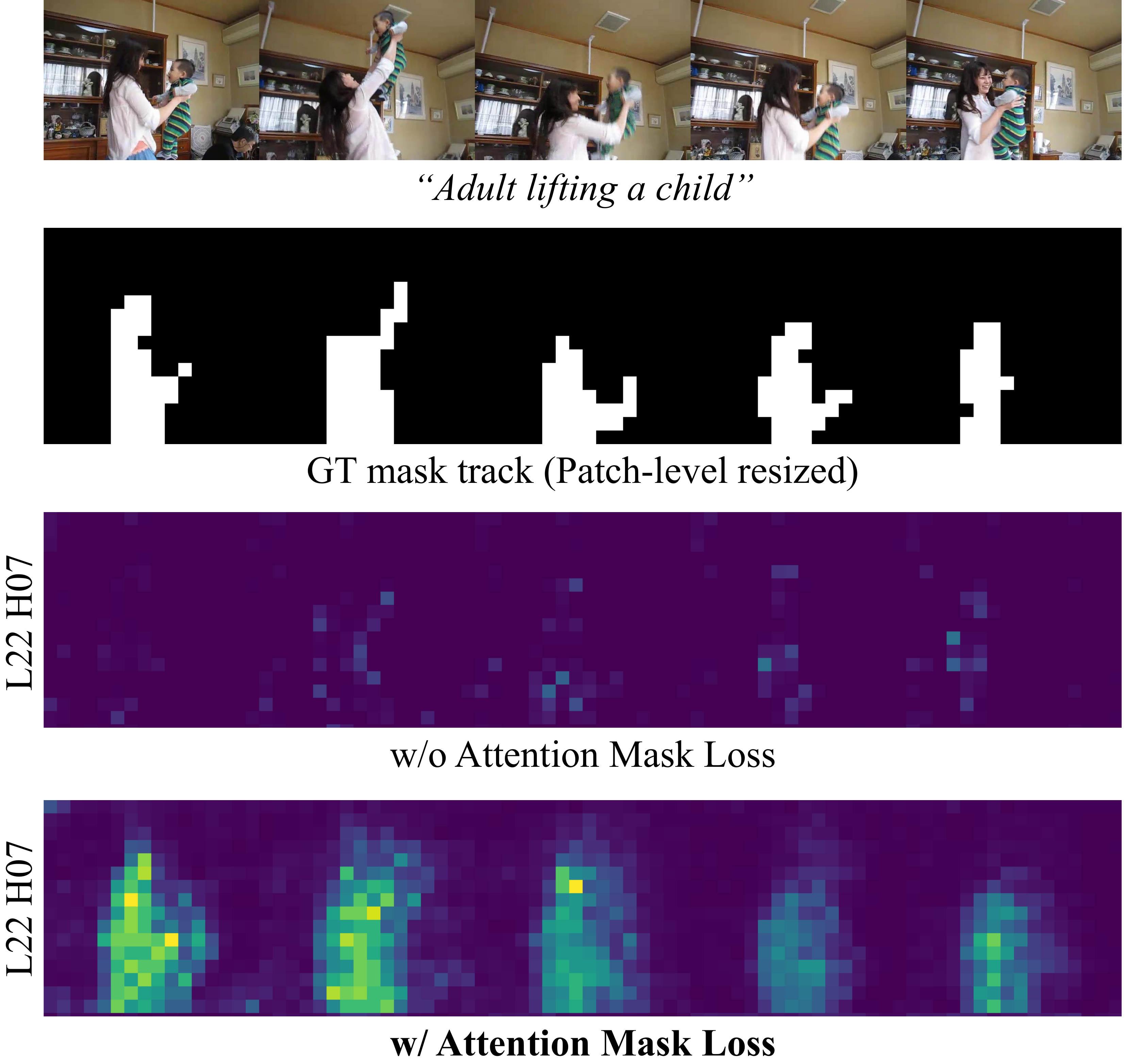}
    \caption{
    \textbf{Effectiveness of our proposed AML.} L22H07 denotes the 7th head of the 22nd layer (indices start at 0).}
    \label{fig:method_attn_mag}
\end{figure}

\begin{table*}[!t]
    \centering
    \begin{tabular}{lccccccccc}
    \toprule
    \multirow{2}{*}{Methods} & \multicolumn{3}{c}{\ours-Actor} & \multicolumn{3}{c}{\ours-Target} & \multicolumn{3}{c}{RVOS} \\
    \cmidrule(lr){2-4}
    \cmidrule(lr){5-7}
    \cmidrule(lr){8-10}
     & \( \mathcal{J} \) & \( \mathcal{F} \) & \( \mathcal{J} \)\&\( \mathcal{F} \) & \( \mathcal{J} \) & \( \mathcal{F} \) & \( \mathcal{J} \)\&\( \mathcal{F} \) & \( \mathcal{J} \) & \( \mathcal{F} \) & \( \mathcal{J} \)\&\( \mathcal{F} \) \\
    \midrule
    Referformer~\cite{wu2022language} & 59.1 & 59.9 & 59.5 & - & - & - & 52.0 & 53.2 & 52.6 \\
    LMPM~\cite{ding2023mevis} & 51.1 & 54.1 & 52.6 & - & - & - & 45.1 & 48.3 & 46.7 \\
    VISA-7B~\cite{yan2024visa} & 57.8 & 57.6 & 57.7 & - & - & - & 49.2 & 50.4 & 49.8 \\
    VideoLISA-3.8B~\cite{bai2024one} & 68.4 & 68.0 & 68.2 & - & - & - & \underline{61.5} & 61.9 & 61.7 \\

    Sa2VA-1B~\cite{yuan2025sa2va} & 69.9 & 72.6 & 71.3 & - & - & -  & 55.4 & 58.7 & 57.0\\
    Sa2VA-4B~\cite{yuan2025sa2va} & 69.6 & 72.3 & 71.0 & - & - & -  & 58.1 & 61.0 & 59.5\\
    
    \midrule
    
    \textbf{\arch-1B} & \underline{71.8} & \underline{74.7} & \underline{73.3} & \underline{65.9} & \underline{68.9} & \underline{67.4} & 60.2 & \underline{63.8} & \underline{62.0} \\
    \textbf{\arch-4B} & \textbf{73.2} & \textbf{75.8} & \textbf{74.5} & \textbf{67.1} & \textbf{69.5} & \textbf{68.3} & \textbf{63.0} & \textbf{66.1} & \textbf{64.5} \\

    \bottomrule
    \end{tabular}
    \caption{\textbf{Quantitative results on \dataset dataset.}
    \arch achieves the highest performance on interaction-aware settings (\ours-Actor and \ours-Target), demonstrating its effectiveness in modeling inter-object dynamics. Notably, the surpassing performance on \ours-Actor indicates that explicitly segmenting both the actor and the target enhances the model's ability to localize the actor itself, reflecting a better understanding of the overall temporal dynamics.
    The best-performing results are presented in \textbf{bold}, while the second-best results are \underline{underlined}.
    }
    \label{tab:main_quan}
\end{table*}

\subsection{Attention mask loss}
During the generation of special tokens, the MLLM produces self-attention score matrices at each transformer layer and head. Each attention map is of size $(N_v + N_t) \times (N_v + N_t)$, where $N_v = T' \times P \times P$ denotes the number of vision tokens and $N_t$ is the number of text tokens. Here, $P \times P$ represents the number of patches per frame.
From each attention map, we extract the attention scores from the special segmentation token (i.e., the query tokens \actor or \target) to all visual tokens. These weights are then reshaped into a spatio-temporal attention map $A^{(l,h)} \in [0,1]^{T' \times P \times P}$ for each layer $l$ and head $h$, aligning with the patch layout of the input video frames.
Notably, we observed that specific layers in the MLLM attend more strongly to visual tokens, indicating better spatial localization potential. However, attention maps from MLLMs are often coarse and do not focus precisely on the object the model aims to segment. To guide these maps toward spatially accurate regions, we introduce attention mask loss (AML), which the brief concept illustrated in Figure~\ref{fig:method_attn_mag}.

We first identify a set of specific layer-head pairs $\mathcal{H}$ using a selection protocol based on vision attention. For each selected $(l,h) \in \mathcal{H}$, we supervise the attention map $A^{(l,h)}$ using the ground-truth binary mask $\mathcal{M'} \in \{0,1\}^{T' \times H \times W}$, which is resized to the patch resolution, resulting in $\mathcal{G}' \in \{0,1\}^{T' \times P \times P}$.
Since our method distinguishes between actor and target objects, we apply supervision to each type jointly. Specifically, the AML is defined as:
\begin{equation}
\mathcal{L}_{\mathrm{AML}} = \sum_{r \in \{\mathrm{act}, \mathrm{tar}\}} \sum_{(l,h) \in \mathcal{H}} \mathrm{BCE}\left(A_r^{(l,h)}, \mathcal{G}'_r\right).
\end{equation}
By explicitly supervising the attention scores to align with the object mask, AML encourages the MLLM to ground special tokens more precisely in the visual domain. This auxiliary loss is jointly optimized with the segmentation loss during training.

\subsection{Overall training loss}
We apply standard pixel-wise cross-entropy loss and dice loss between the predicted mask $\hat{\mathcal{M}}$ and ground-truth mask track $\mathcal{M}$:
\begin{equation}
    \mathcal{L}_{\mathrm{seg}} = \sum_{r \in \{\mathrm{act, tar}\}} {\mathcal{L}_{\mathrm{CE}}(\hat{\mathcal{M}}_{\mathrm{r}}, \mathcal{M}_{\mathrm{r}}) + \mathcal{L}_{\mathrm{Dice}}(\hat{\mathcal{M}}_{\mathrm{r}}, \mathcal{M}_{\mathrm{r}})}.
\end{equation}
When the referring expression describes an interaction, the segmentation loss is computed for both masks, $\hat{\mathcal{M}}_{\mathrm{act}}$ and $\hat{\mathcal{M}}_{\mathrm{tar}}$. Otherwise, it is computed only on $\hat{\mathcal{M}}_{\mathrm{act}}$.
Additionally, we include a text loss $\mathcal{L}_{\mathrm{text}}$, defined as the cross-entropy loss over the predicted and ground-truth referring expressions.
Consequently, the total training loss is defined as:
\begin{equation}
\mathcal{L}_{\mathrm{total}} = \mathcal{L}_{\mathrm{seg}} + \lambda_{\mathrm{AML}} \cdot \mathcal{L}_{\mathrm{AML}} + \lambda_{\mathrm{text}} \cdot \mathcal{L}_{\mathrm{text}},
\end{equation}
where $\lambda_{\mathrm{AML}}$ and $\lambda_{\mathrm{text}}$ are weighting coefficients for the attention mask loss and text loss, respectively.

\section{Experiments}

In this section, we present experimental results to evaluate the effectiveness of our proposed approach, \arch. We report performance on the \dataset evaluation set compared to various baselines, and further analyze \arch through ablation studies.
All experiments follow standard RVOS metrics~\cite{khoreva2019video, seo2020urvos, ding2023mevis}, using the average of region similarity $\mathcal{J}$ and contour accuracy $\mathcal{F}$, denoted as $\mathcal{J}\&\mathcal{F}$.
Further experimental details are provided in the Appendix, including implementation details (Appendix~\ref{suppl:implementation}), an extended analysis of the proposed AML (Appendix~\ref{suppl:AML_analysis}), as well as quantitative and zero-shot evaluations on multiple RVOS benchmarks, together with additional qualitative results (Appendix~\ref{suppl:experiment}).

\subsection{Experimental results}

\begin{figure}[!t]
    \centering
    \includegraphics[width=\columnwidth]{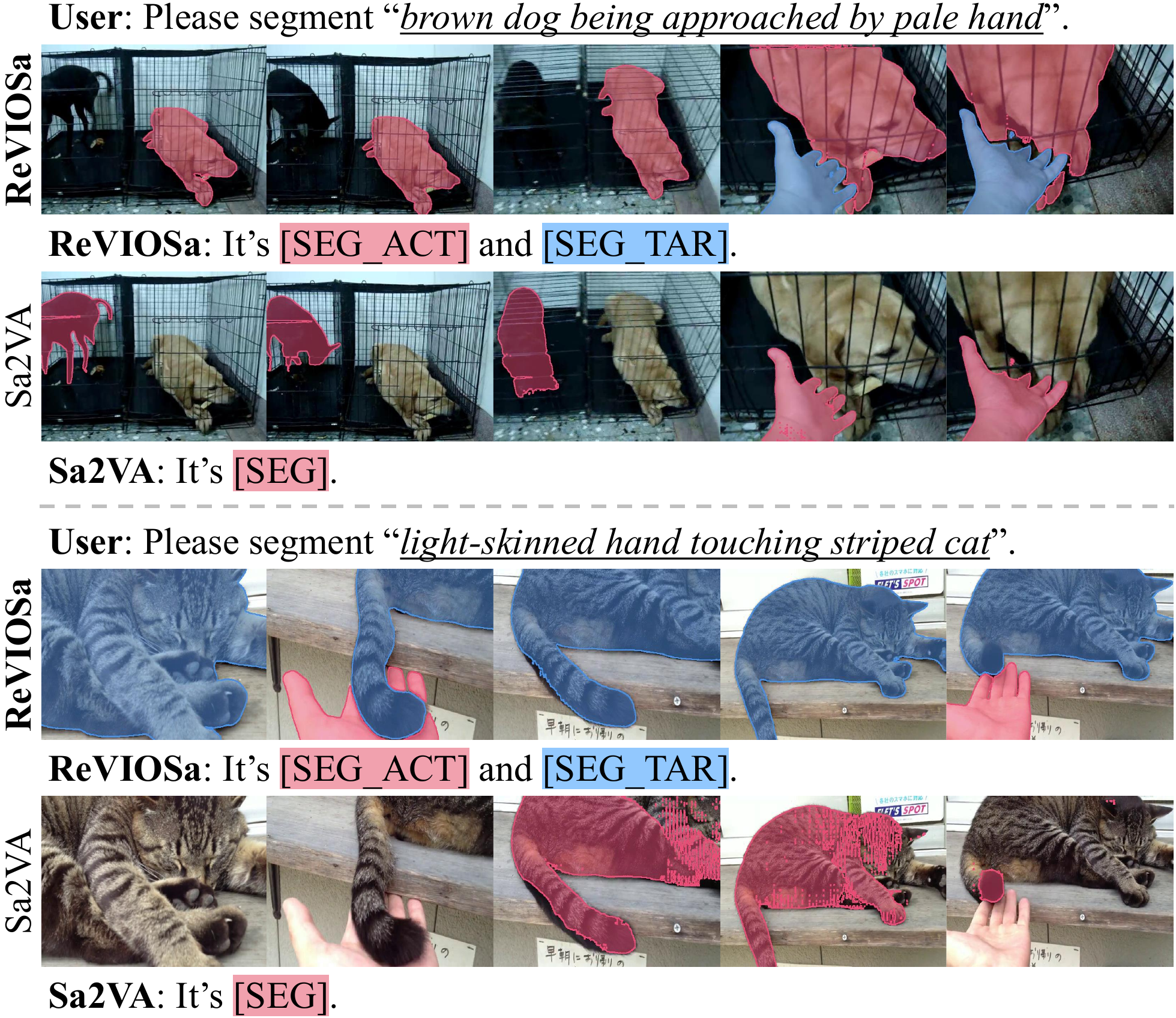}
    \caption{\textbf{Qualitative results.}
    Compared to the previous RVOS method,  Sa2VA~\citep{yuan2025sa2va}, \arch accurately segments both the actor and the target objects when given an interaction expression, demonstrating its ability to distinguish object roles.
    }
    \label{fig:main_qual}
\end{figure}

\paragraph{Quantitative results.}
Table~\ref{tab:main_quan} presents quantitative results under three evaluation settings: \ours-Actor, \ours-Target, and RVOS. The first two are newly introduced protocol to evaluate \ours, which focuses on role-specific segmentation of actors and targets for each interaction expression sample. RVOS represents the standard RVOS setting which only segments the actor objects, which is conducted for all expression samples.
Importantly, previous RVOS approaches~\cite{wu2022language, ding2023mevis, yan2024visa, yuan2025sa2va} are designed to segment only the actor, and thus are not applicable to the \ours-Target setting, highlighting the novelty and necessity of our proposed task.
Even so, our proposed \arch demonstrates competitive performance on both the \ours-Actor and RVOS. The 1B model already surpasses previous methods on most metrics, while the 4B model achieves state-of-the-art performance across all metrics.
This demonstrates that the capability of accurate distinction of the roles of actor and target enhances the model’s capability to capture the overall object behavior and complex spatiotemporal inter-object relationships.
Additionally, we report the performance of \arch on standard RVOS benchmarks in Appendix~\ref{suppl:quan}, along with comparisons of training datasets on the MeVIS benchmark (Appendix~\ref{suppl:mevis_quan}), zero-shot evaluations across multiple RVOS benchmarks (Appendix~\ref{suppl:zeroshot}).

\begin{figure}[!t]
    \centering
    \includegraphics[width=\columnwidth]{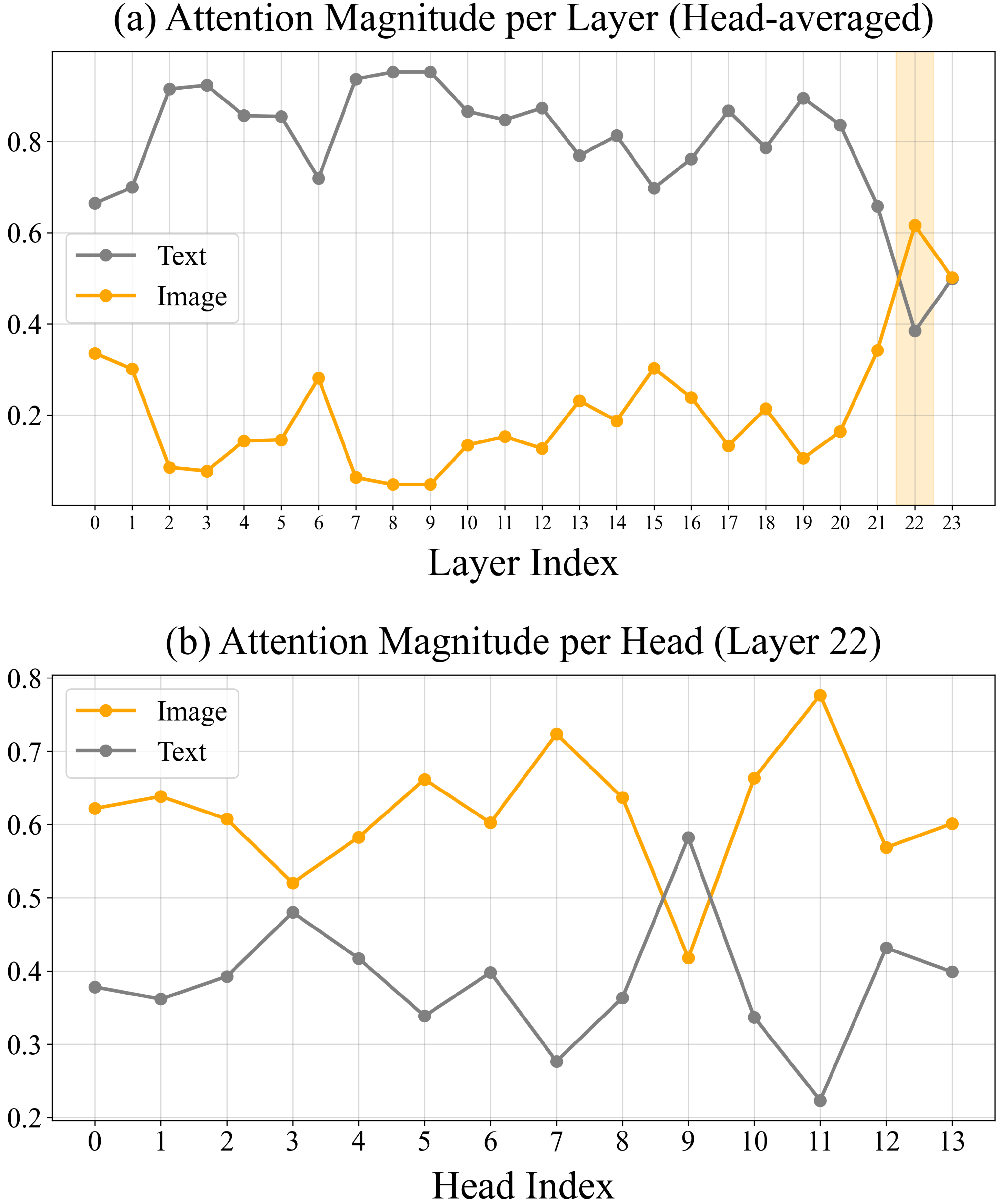}
    \caption{
    \textbf{Attention magnitude across layers and heads.}
    }
    \label{fig:selection}
\end{figure}

\begin{table}[!t]
    \centering
    \setlength{\tabcolsep}{1.3mm}
    \begin{tabular}{c|ccccc}
    \toprule
    \multirow{2}{*}{All Heads} & $k=1$ & $k=2$ & $k=3$ & $k=4$ & $k=5$ \\
    & (H11) & (+ H07) & (+ H10) & (+ H05) & (+ H08) \\
    \midrule
    60.7 & 61.3 & 60.5 & 61.2 & \textbf{62.0} & 60.6 \\
    \bottomrule
    \end{tabular}
    \caption{\textbf{Performance comparison of AML applied to top-$k$ attention heads in layer 22.}
    Empirically, applying AML to the top-4 heads yields the best performance.
    }
    \label{tab:head_selection}
\end{table}

\paragraph{Qualitative results.}
Figure~\ref{fig:main_qual} compares qualitative results under the \ours setting, where both the input video and expression involve multiple interacting objects. In these complex cases, the previous RVOS approach Sa2VA~\cite{yuan2025sa2va}, fails to identify the referred object under interaction. In contrast, \arch is explicitly trained to distinguish object roles, enabling more precise recognition and segmentation of both actor and target objects. Additional qualitative results can be found in Appendix~\ref{suppl:qual}.

\subsection{Analysis}
In this section, we analyze two key aspects of our proposed architecture. First, we investigate a layer-head selection strategy for applying attention mask loss (AML), selecting the most effective layers and heads for supervision. Second, we conduct an ablation study to evaluate the effectiveness of interaction-aware special tokens and AML. All experiments are conducted using the \arch-1B model.
Additional results and analysis, including for \arch-4B, are provided in the Appendix~\ref{suppl:AML_analysis}, covering the motivation of AML (Appendix~\ref{suppl:motivation}), details of layer–head selection for AML (Appendix~\ref{suppl:selection}), attention comparison between interaction-aware special tokens (Appendix~\ref{suppl:attn_act_tar}), and attention visualization and analysis (Appendix~\ref{suppl:attn_vis}).

\begin{figure}[!t]
    \centering
    \includegraphics[width=\columnwidth]{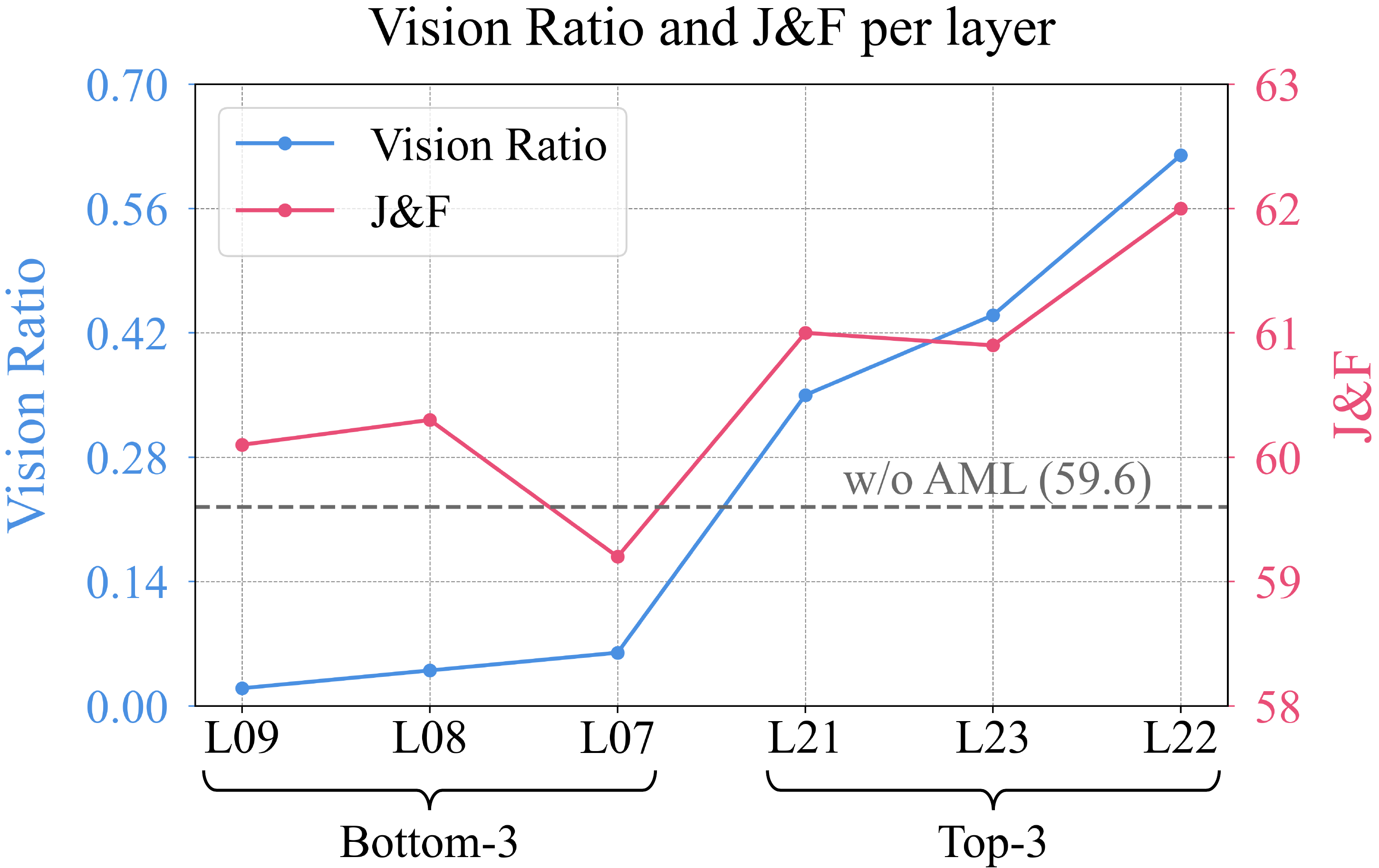}
    \caption{
    \textbf{Layer-wise impact of attention mask loss (AML).}
    Layers with stronger vision focus ( L22, L23, L21) exhibit greater improvements in {$\mathcal{J}\&\mathcal{F}$} compared to layers with weaker vision focus (L09, L08, L07).
    }
    \label{fig:top_bot_layer}
\end{figure}
\begin{table}[!t]
    \centering
        \begin{tabular}{c|c|c|ccc}
            \toprule
            & \actor & \multirow{2}{*}{AML} & \multirow{2}{*}{$\mathcal{J}$} & \multirow{2}{*}{$\mathcal{F}$} & \multirow{2}{*}{$\mathcal{J}\&\mathcal{F}$} \\
            & \target &  &  &  &  \\
            \midrule
            (i) & \xmark & \xmark & 55.4 & 58.7 & 57.0 \\
            (ii) & \xmark & \cmark & 57.4 & 59.6 & 58.5 \\
            (iii) & \cmark & \xmark & 57.8 & 61.3 & 59.6 \\
            (iv) & \cmark & \cmark & \textbf{60.2} & \textbf{63.8} & \textbf{62.0} \\
            \bottomrule
        \end{tabular}
    \caption{\textbf{Ablation study on \arch architecture.}
    Both (ii) AML and (iii) the interaction-aware special tokens contribute significant performance gains over (i) the baseline. Our full model (iv) \arch achieves the highest performance among all configurations.
    }
    \label{tab:ablation}\vspace{-10pt}
\end{table}

\paragraph{Layer-head selection for AML.}
We analyze layer-head configurations for applying attention mask loss (AML), which supervises attention maps to enhance the focus on relevant object regions. To identify suitable configurations, we investigate the attention map across layers and heads, where the query is \actor token and the keys are vision tokens.

As shown in Figure~\ref{fig:selection}(a), Layer 22 exhibits the highest head-averaged attention to vision tokens, making it the most suitable layer for AML application. We then analyze the head-wise attention scores within Layer 22 (Figure~\ref{fig:selection}(b)) and empirically find that applying AML to the top-4 heads yields the best performance, as reported in Table~\ref{tab:head_selection}.

Based on these findings, we adopt the following strategy: (i) select the layer with the highest head-averaged attention to vision tokens, and (ii) apply AML to its top-4 heads only.
To further validate this selection strategy, we compare AML applied to the top-3 versus bottom-3 layers. As shown in Figure~\ref{fig:top_bot_layer}, the top-3 layers (L22, L23, L21) consistently outperform the bottom-3 (L09, L08, L07), with Layer 22 alone yielding a +2.4 improvement in $\mathcal{J}\&\mathcal{F}$ over the baseline.

These results confirm that supervising attention in vision-focused layers is crucial for performance, and demonstrate the effectiveness of AML as a training signal.

\paragraph{Ablation studies.}
We perform an ablation study to assess the individual and combined contributions of two core components: the interaction-aware special tokens (\texttt{\actor} and \texttt{\target}) and the proposed attention mask loss (AML).
As presented in Table~\ref{tab:ablation}, each component independently improves model performance over the baseline (57.0 $ \mathcal{J}\&\mathcal{F}$), with AML contributing +1.5 and interaction-aware tokens adding +2.6. 
When both are used together, the model achieves a performance of 62.0 on the \dataset evaluation set, demonstrating their complementary benefits in understanding and segmenting interacting objects.

\section{Conclusion}

We present \ours, a novel task that extends the standard RVOS by requiring the segmentation of both actor and target objects from a single interaction expression, thereby explicitly modeling inter-object dynamics.
To support this, we present \dataset, a large-scale dataset with over 127K expressions and distinct actor-target annotations for interaction expressions.
We also propose \arch, an MLLM-based model with interaction-aware tokens and attention mask loss for precise role-specific segmentation.
Extensive experiments validate the effectiveness of modeling interaction, with \arch achieving state-of-the-art performance on \dataset.

\clearpage
\newpage

\twocolumn[{
\begin{center}
    \LARGE \textbf{InterRVOS: Interaction-Aware Referring Video Object Segmentation}\\[0.5em]
    \Large \textbf{-- Appendix --}\\[1.5em]
\end{center}
}]

\appendix
\setcounter{secnumdepth}{3}
\setcounter{section}{0}
\setcounter{figure}{0}
\setcounter{table}{0}

\renewcommand{\thesection}{\Alph{section}}
\renewcommand{\thefigure}{A\arabic{figure}}
\renewcommand{\thetable}{A\arabic{table}}

In the appendix, we provide additional details and analyses that further support the results and findings presented in the main paper.
First, Section~\ref{suppl:implementation} outlines implementation details, including model configurations and training settings.
Second, Section~\ref{suppl:AML_analysis} presents additional analyses on attention mask loss (AML), covering the motivation behind AML in Section~\ref{suppl:motivation} and the detailed layer-head selection strategy for both the 1B and 4B models in Section~\ref{suppl:selection}.
In addition, Sections~\ref{suppl:attn_act_tar} and~\ref{suppl:attn_vis} analyze the attention maps of \arch, demonstrating its ability to model interactions effectively.
Further experimental results regarding the effectiveness of both \arch and \dataset are provided in Section~\ref{suppl:experiment}, including quantitative evaluations on RVOS benchmarks (Section~\ref{suppl:quan}), comparison of training datasets on the MeVIS benchmark (Section~\ref{suppl:mevis_quan}), zero-shot evaluation across various training datasets on multiple RVOS benchmarks (Section~\ref{suppl:zeroshot}), and additional qualitative examples (Section~\ref{suppl:qual}).
Finally, Section~\ref{suppl:dataset_detail} provides additional details on \dataset, including our data annotation pipeline (Section~\ref{suppl:annotation_pipeline}), additional examples of \dataset (Section~\ref{suppl:data_samples}), the video clip extraction procedure describing how the training and evaluation sets were derived from source videos (Section~\ref{suppl:video_clip}), and overall dataset statistics (Section~\ref{suppl:statistics}).

\section{Implementation details}
\label{suppl:implementation}
For the proposed architecture \arch, we utilize InternVL-2.5~\cite{chen2024expanding} as the base model for multimodal large language model (MLLM), applying LoRA~\cite{hu2022lora} tuning exclusively. For the segmentation module, we adopt SAM2~\cite{ravi2024sam} and fine-tune only its decoder while keeping the image encoder frozen. The model is trained for 10 epochs with a batch size of 2. We report results using two model scales: 1B and 4B. The 1B model is trained on 4 NVIDIA RTX 3090 GPUs for 12 hours, whereas the 4B model is trained on 4 NVIDIA A6000 GPUs for 16 hours.

\section{Analysis on AML}
\label{suppl:AML_analysis}
\begin{figure}[!t]
    \centering
    \includegraphics[width=\columnwidth]{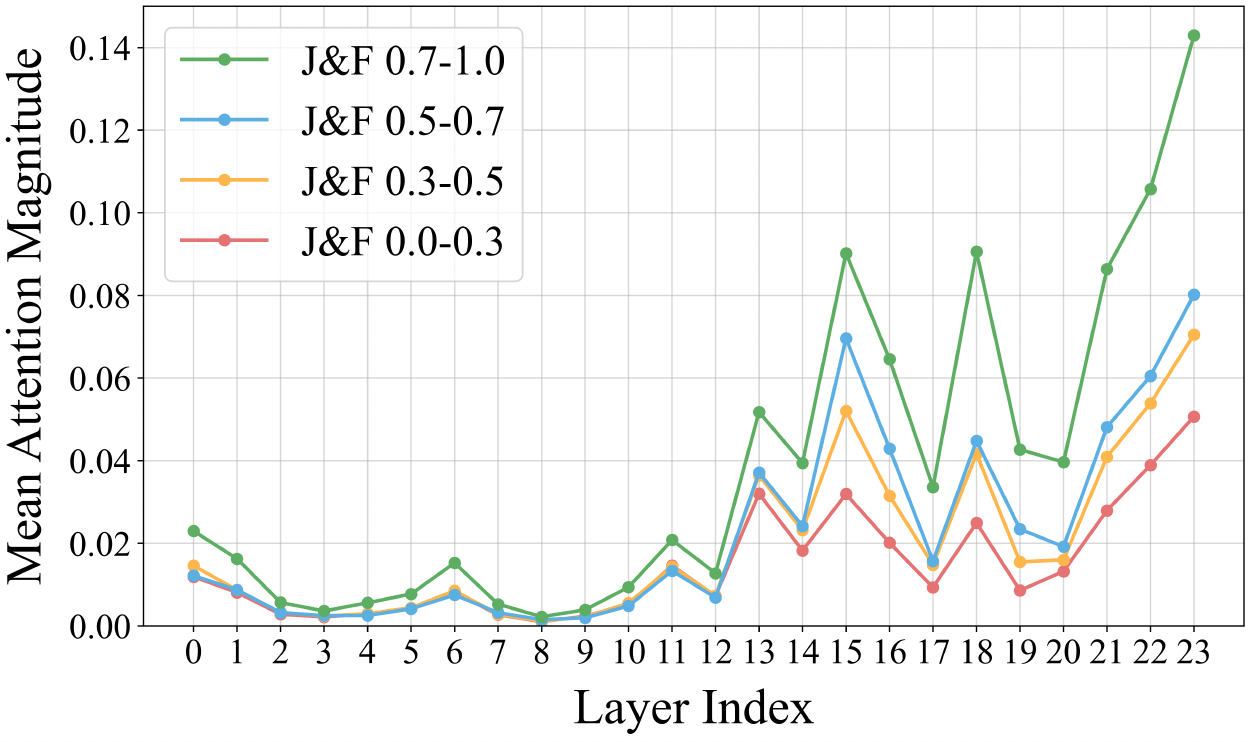}
    \caption{\textbf{Correlation between attention scores and segmentation quality.} 
    For each $\mathcal{J}\&\mathcal{F}$ score interval, we plot the mean attention score within the ground-truth mask region, averaged across heads for each layer. Higher-performing samples consistently exhibit greater attention within the mask regions, motivating the use of attention mask supervision.
    }
    \label{fig:suppl_motivation}
\end{figure}

In this section, we present our analysis of the attention maps from the MLLM and the detailed layer-head selection process for both the 1B and 4B models.
Specifically, Section~\ref{suppl:motivation} provides the motivation for applying attention mask loss (AML) by examining the correlation between attention aggregation and segmentation performance.
Section~\ref{suppl:selection} then describes how we select appropriate layer-head pairs based on attention magnitude to vision tokens, and describes the detailed selection strategies across model scales (1B and 4B).

\begin{figure*}[!ht]
    \centering
    \includegraphics[width=0.8\textwidth]{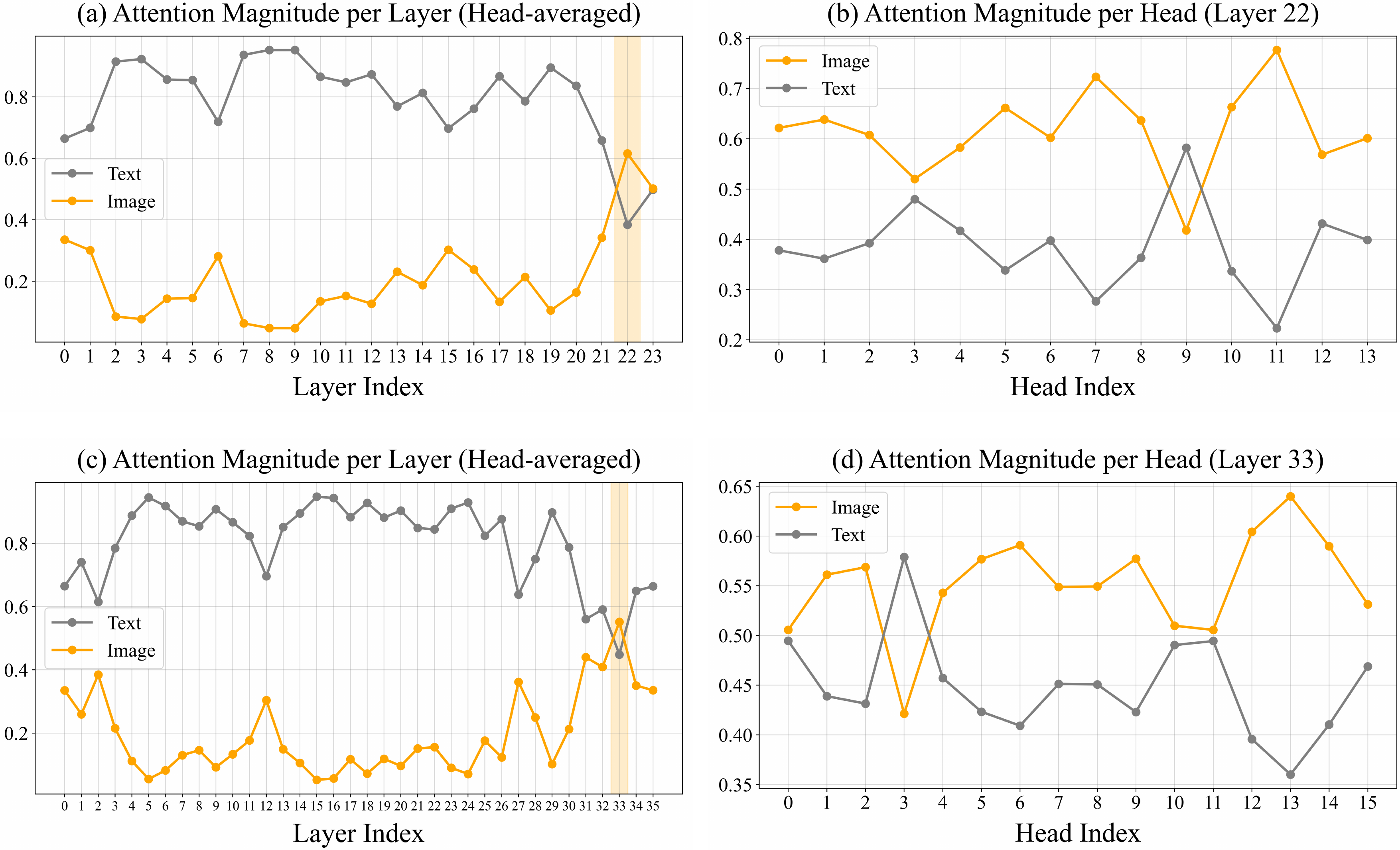}
    \caption{\textbf{Attention magnitude across layers and heads.}
    Each figure illustrates the attention magnitude from the special token (query) to vision tokens (key) across different layers and heads of the MLLM.
    (a) Layer-wise head-averaged attention scores for the 1B model.  
    (b) Head-wise attention scores within Layer 22 of the 1B model.  
    (c) Layer-wise head-averaged attention scores for the 4B model.  
    (d) Head-wise attention scores within Layer 33 of the 4B model.  
    These results guide the selection of the top-1 layer and its top-4 heads for AML supervision.}
    \label{fig:suppl_selection}
\end{figure*}

\begin{figure}[!ht]
    \centering
    \includegraphics[width=0.8\columnwidth]{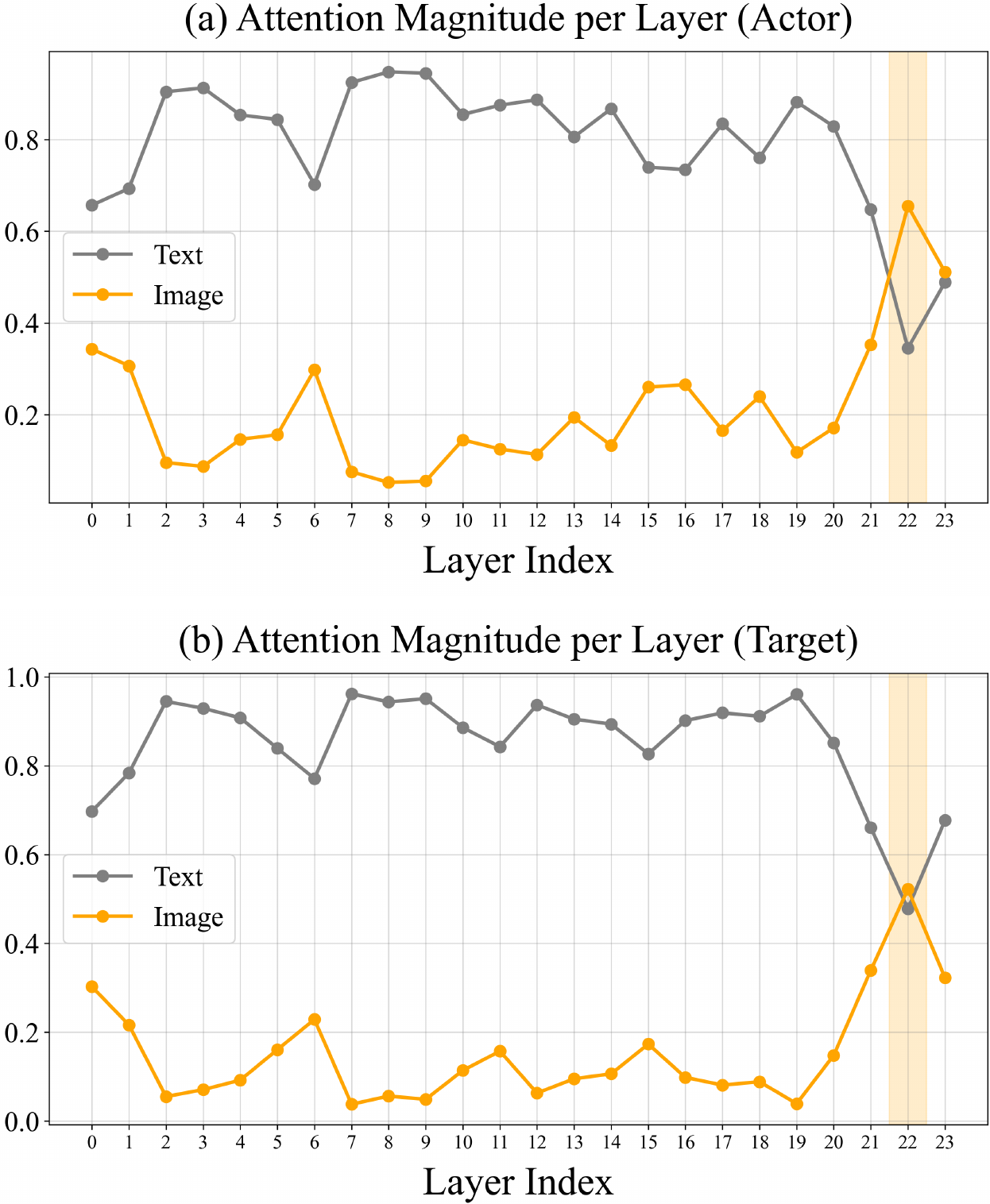}
    \caption{\textbf{Comparison between attention magnitude of interaction-aware special tokens.}
    }
    \label{fig:suppl_act_tar}
\end{figure}

\subsection{Motivation of AML}
\label{suppl:motivation}
While our main paper demonstrates that certain layers and heads in the MLLM exhibit stronger attention to vision tokens, suggesting their potential for object-level localization, this alone does not justify the need for explicit supervision on the attention maps. To further motivate the introduction of attention mask loss (AML), we analyze how well the attention maps from special tokens (i.e., \actor and \target) align with the actual segmentation object regions.

Specifically, we compute the sum of attention scores over the ground-truth mask regions, which is the cumulative attention weight assigned by each special token (query) to the visual tokens (key) corresponding to the object. This analysis extends the layer-wise, head-averaged attention score evaluation presented in the main paper by directly quantifying the spatial correspondence between attention and object masks.

To validate this correlation, we group the samples into four intervals based on their segmentation performance (i.e., $\mathcal{J}\&\mathcal{F}$ scores of 0–0.3, 0.3–0.5, 0.5–0.7, and 0.7–1.0) and plot the average attention score within the mask region for each layer across all heads (Figure~\ref{fig:suppl_motivation}). As shown, samples with higher $\mathcal{J}\&\mathcal{F}$ scores exhibit consistently higher attention concentration in the ground-truth mask regions. This trend suggests that stronger alignment between the attention maps and the segmentation masks is associated with better segmentation outcomes.

This empirical observation supports the need to explicitly guide the attention maps to focus on the correct object regions. Based on this insight, we apply a binary cross-entropy loss between the attention maps and the resized ground-truth masks, supervising only the selected layer-head pairs identified through our layer-head selection strategy. This attention mask loss directly encourages the model to ground the \actor and \target tokens more precisely in the object region, thereby improving the downstream segmentation performance.

\begin{figure*}[!ht]
    \centering
    \includegraphics[width=0.75\textwidth]{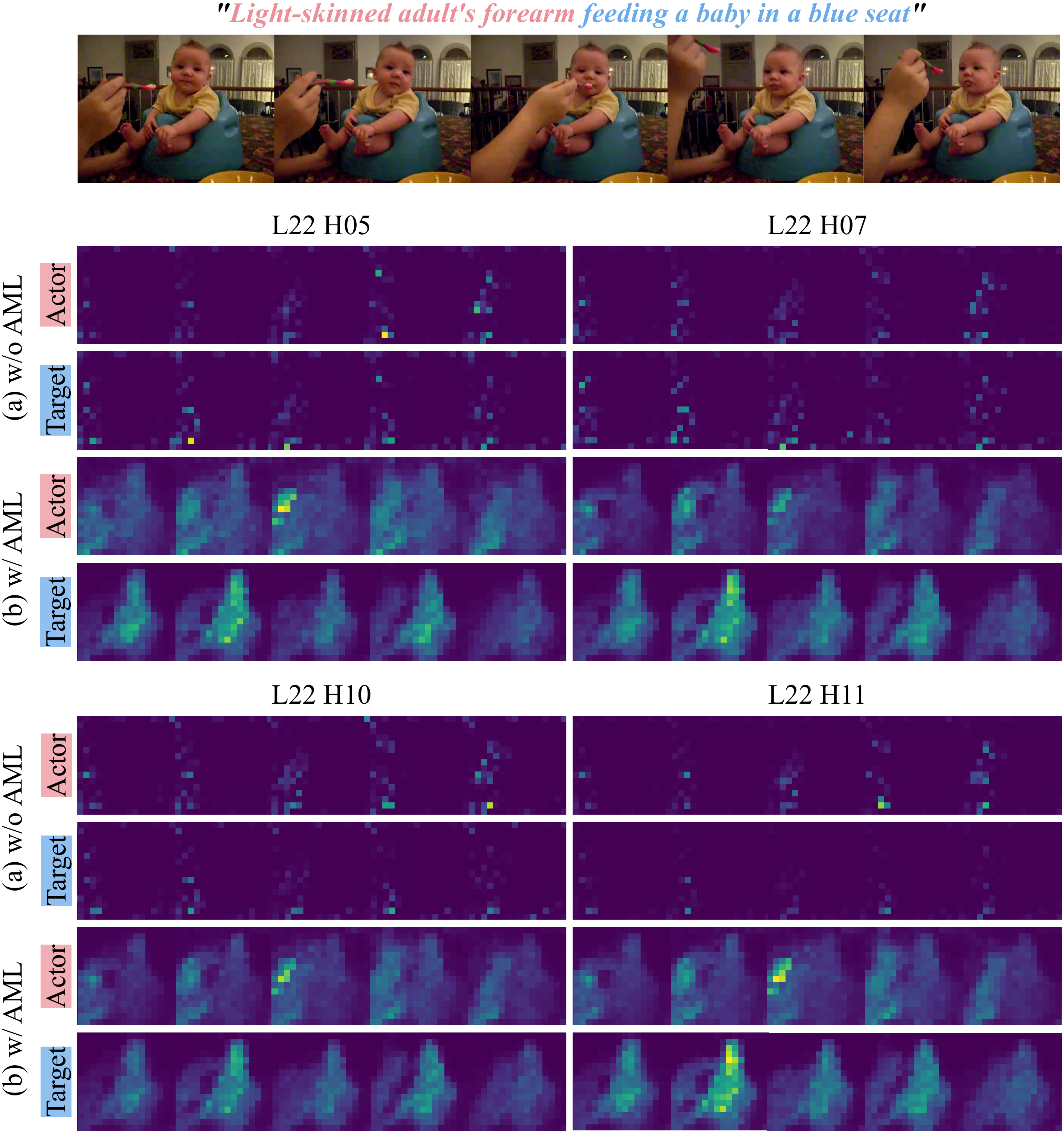}
    \caption{\textbf{Visualization of attention maps for trained layer-head pairs.}
    Comparison between attention maps of baseline and w/AML.
    }
    \label{fig:suppl_attn_vis}
\end{figure*}

\begin{figure*}[!ht]
    \centering
    \includegraphics[width=0.75\textwidth]{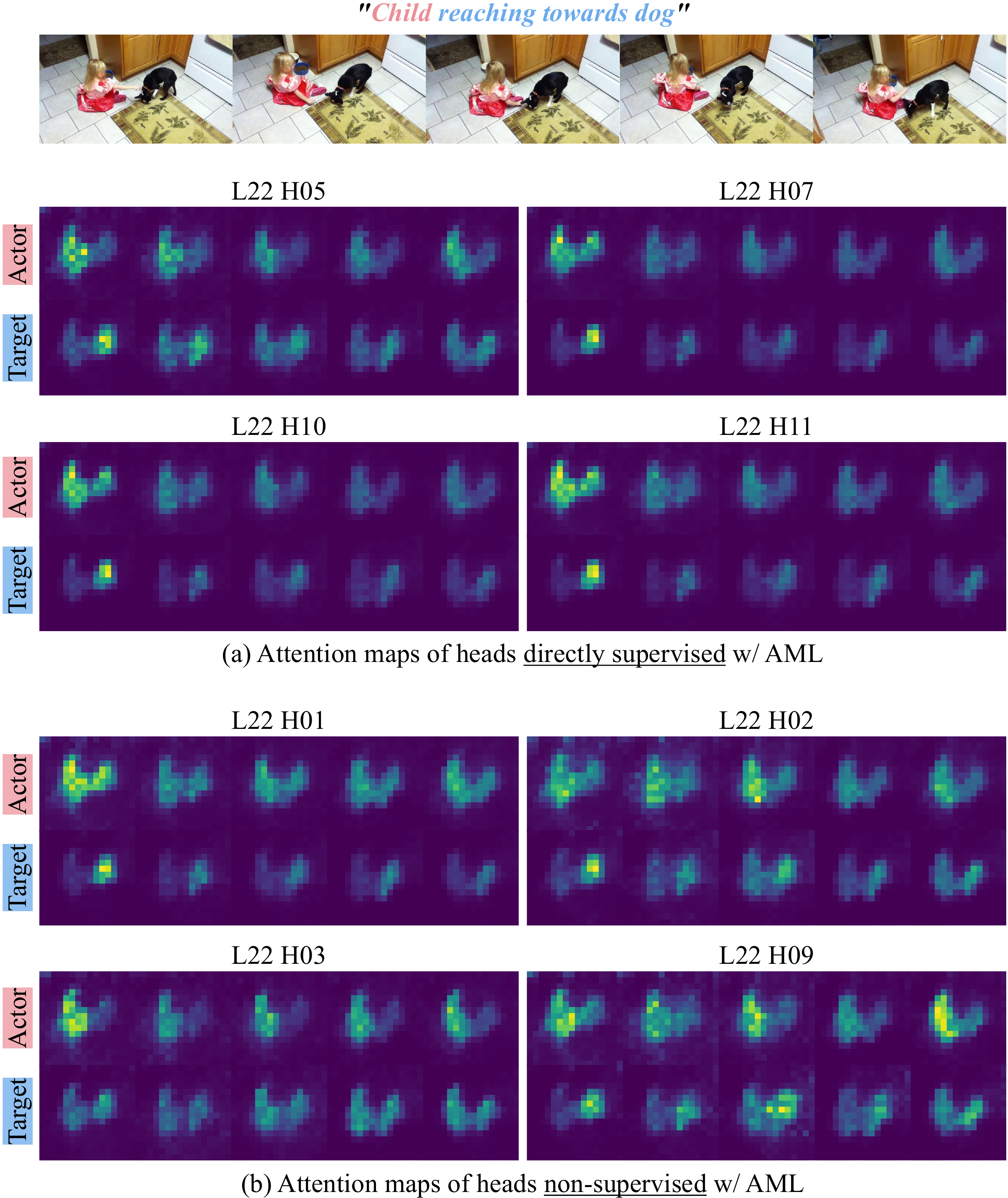}
    \caption{\textbf{Effect of AML on supervised vs. non-supervised heads.}  
    Attention maps from various heads of Layer 22 of the 1B model.
    (a) Heads directly supervised by AML.  
    (b) Non-supervised heads within the same layer.  
    Even without direct supervision, non-supervised heads show improved focus, indicating that AML induces a beneficial effect to nearby layers and heads.
    }
    \label{fig:suppl_attn_diff}
\end{figure*}

\subsection{Layer-head selection for AML}
\label{suppl:selection}
As described in the main paper, our layer-head selection strategy for AML first identifies the layer with the highest head-averaged attention to vision tokens, and then selects the top-4 heads within that layer.
The attention scores across layers and heads for both 1B and 4B models are visualized in Figure~\ref{fig:suppl_selection}, where (a) and (b) correspond to the 1B model, and (c) and (d) to the 4B model.
For the 4B model, Figure~\ref{fig:suppl_selection}(c) shows that Layer 33 exhibits the highest average attention to vision tokens. Within this layer, we further analyze the head-wise attention scores (Figure~\ref{fig:suppl_selection}(d)) and select the top-4 heads, H13, H12, H14, and H06, for AML supervision.

By consistently applying AML to layers and heads with strong attention to vision tokens, we effectively deliver spatial supervision to the most responsive components of the MLLM.

\subsection{Attention comparison between interaction-aware special tokens}
\label{suppl:attn_act_tar}
We compare the attention magnitude of the interaction-aware special tokens, \actor and \target, to examine whether separate selection strategies are necessary. In the main paper, all layer-head selections for AML were conducted based on the attention maps of the \actor token. This decision is justified by the observation that the attention magnitudes and distributions of \actor and \target tokens are similar.

As shown in Figure~\ref{fig:suppl_act_tar}, (a) presents the attention magnitude from the \actor token to vision tokens, while (b) shows the corresponding scores for the \target token. Notably, both tokens exhibit the highest vision attention at Layer 22, indicating that the same top-1 vision-attending layer is shared across the two roles. The distributions are closely aligned across layers, indicating that applying the selection strategy based solely on the \actor token is sufficient for effective supervision of both roles.

\subsection{Attention visualization}
\label{suppl:attn_vis}

\paragraph{Comparison of attention maps with and without AML.}
Figure~\ref{fig:suppl_attn_vis} illustrates the differences in attention maps between the models trained without AML and with AML, denoted as w/o AML and w/ AML, respectively. We visualize the attention maps from the 1B model, focusing on the specific layer-head pairs where AML supervision was applied.
Without AML, the attention maps are notably sparse and diffuse, showing limited focus on the relevant object regions. In contrast, with AML, the attention becomes significantly sharper and more concentrated within the correct object areas. This is evident for both the \actor (adult’s hand) and \target (child) tokens, each token attending reliably to the object it is responsible for segmenting.
These results demonstrate that AML enhances the MLLM’s ability to allocate attention \textit{distinctly} for each interaction-aware token, thereby enabling role-specific segmentation.

\paragraph{Effect of AML on supervised and non-supervised heads.}
Figure~\ref{fig:suppl_attn_diff} compares the attention maps from (a) heads directly supervised by AML and (b) non-supervised heads within the same layer (Layer 22 of the 1B model). The supervised heads correspond to those explicitly selected for AML training, while the non-supervised heads did not receive direct supervision.
Notably, we observe that the non-supervised heads also exhibit improved attention focus on the target object regions, despite not being explicitly trained with AML. This indicates that the supervision signal from AML can propagate within a layer, positively influencing other heads and contributing to more consistent spatial grounding across the entire attention module.

\begin{table*}[!t]
    \centering
    \begin{tabular}{l|ccc}
    \toprule
    Methods & MeViS & Ref-Youtube-VOS & Ref-DAVIS \\
    \midrule
    LISA-7B~\cite{lai2024lisa} & 39.4 & 54.3 & 64.8 \\
    LISA-13B~\cite{lai2024lisa} & 37.9 & 54.4 & 66.0 \\
    TrackGPT-7B~\cite{zhu2023tracking} & 40.1 & 56.4 & 63.2 \\
    TrackGPT-13B~\cite{zhu2023tracking} & 41.2 & 59.5 & 66.5 \\
    VISA-7B~\cite{yan2024visa} & 43.5 & 61.5 & 69.4 \\
    VISA-13B~\cite{yan2024visa} & 44.5 & 63.0 & 70.4 \\
    Sa2VA-4B~\cite{yuan2025sa2va} & \underline{46.2} &\underline{70.0} & \textbf{73.8} \\
    
    \midrule
    
    \textbf{\arch-4B} & \textbf{49.3} & \textbf{70.5} & \underline{71.6} \\

    \bottomrule
    \end{tabular}
    \caption{\textbf{Quantitative results on RVOS benchmarks.}
    }
    \label{tab:suppl_quan}
\end{table*}

\begin{table*}[!t]
    \centering
    \begin{tabular}{c|c|c|cc}
    \toprule
    \multirow{2}{*}{Dataset} & \multirow{2}{*}{Setting} & Ref-SAV & \ours-28K & \ours-71K \\
     &  & (Videos 37k / Exps. 72K) & (Videos 2k / Exps. 28K) & (Videos 5K / Exps. 71K) \\
    \midrule
    \multirow{2}{*}{MeViS valid} & Joint Training & 46.8 & \textbf{48.5} & \underline{47.1} \\
     & Zero-shot & 32.8 & \underline{40.2} & \textbf{41.8} \\
    \midrule
    \multirow{2}{*}{MeViS valid\_u} & Joint Training & 53.0 & \underline{54.6} & \textbf{54.8} \\
     & Zero-shot & 40.1 & \underline{50.1} & \textbf{50.5} \\
    \bottomrule
    \end{tabular}
    \caption{\textbf{Effectiveness of \dataset.}
    Despite using fewer samples, models trained on \ours-28K and \ours-71K outperform the Ref-SAV dataset~\cite{yuan2025sa2va} (72K)  on MeViS~\cite{ding2023mevis} benchmark in both the joint training setting (with MeViS~\cite{ding2023mevis} train set) and the zero-shot setting (with only \ours-28K and \ours-71K train sets). This highlights the superior data efficiency and interaction-centric supervision quality of the \dataset dataset.
    }
    \label{tab:suppl_mevis_joint}
\end{table*}

\section{Additional experimental results}
\label{suppl:experiment}

In this section, we provide further experimental results supporting the impact of our dataset and architecture.

\subsection{Quantitative results on RVOS benchmarks}
\label{suppl:quan}
Table~\ref{tab:suppl_quan} shows that our model \arch performs competitively on standard RVOS benchmarks~\cite{ding2023mevis, seo2020urvos, khoreva2019video}.
Notably, despite its smaller size (4B), \arch outperforms several existing approaches built on larger base models (7B or 13B), highlighting its effectiveness.
While our method leverages interaction-aware special tokens and attention mask loss (AML), the tokens are not compatible with the standard RVOS setting. Thus, we evaluate a variant using only AML for standard RVOS benchmarks.

\begin{table*}[!t]
    \centering
    \begin{tabular}{ccccc}
    \toprule
    \multirow{3}{*}{Dataset} & \multicolumn{3}{c}{Baseline} & \arch \\
    \cmidrule(lr){2-4}
    \cmidrule(lr){5-5}
    & ReVOS & Ref-SAV & \dataset & \dataset \\
    \midrule
    MeViS valid & 39.6 & 32.8 & \underline{40.4} & \textbf{42.4} \\
    MeViS valid\_u & 49.1 & 40.1 & \underline{49.5} & \textbf{50.1} \\
    Ref-Youtube-VOS & 57.5 & 54.2 & \textbf{61.2} & \underline{60.3} \\
    Ref-DAVIS & 62.8 & 62.1 & \underline{65.9} & \textbf{66.2} \\
    \bottomrule
    \end{tabular}
    \caption{\textbf{Zero-shot evaluation on standard RVOS benchmarks.} This table compares the generalization ability of models trained on three datasets by evaluating them in a zero-shot manner on conventional RVOS benchmarks: MeViS~\cite{ding2023mevis}, Ref-Youtube-VOS~\cite{seo2020urvos}, and Ref-DAVIS~\cite{khoreva2019video}. The baseline model is Sa2VA~\cite{yuan2025sa2va}, and \dataset consistently outperforms models trained on other datasets, demonstrating the effectiveness of our interaction-centric data. We also report results from \arch-1B trained on \dataset, which show that even without specific adaptation to interaction-sparse benchmarks, the model maintains competitive performance.}
    
    \label{tab:suppl_zeroshot}
\end{table*}

\subsection{Comparison of training datasets on MeViS benchmark}
\label{suppl:mevis_quan}
Table~\ref{tab:suppl_mevis_joint} compares the performance of the Sa2VA~\cite{yuan2025sa2va} baseline when trained on different datasets and evaluated on the MeViS~\cite{ding2023mevis} benchmark. Although Ref-SAV~\cite{yuan2025sa2va} is a large-scale dataset with 37K videos and 72K expressions, our subset training dataset—with only 2K videos and 28K expressions—achieves better performance. Even when controlling the sample size by maintaining a comparable number of expressions, models trained on our dataset (\ours-71K) outperform those trained on Ref-SAV.
The gap is especially notable in the zero-shot setting, where the model is evaluated on MeViS without having seen any MeViS samples during training. This indicates that Ref-SAV, while large, is limited by its single-object-centric design. In contrast, our dataset, which is automatically constructed to be diverse and interaction-aware, provides more effective supervision for video understanding tasks.

\subsection{Zero-shot evaluation}
\label{suppl:zeroshot}
Table~\ref{tab:suppl_zeroshot} presents zero-shot evaluation results of models trained on different datasets—ReVOS~\cite{yan2024visa}, Ref-SAV~\cite{yuan2025sa2va}, and \dataset—on three standard RVOS benchmarks: MeViS~\cite{ding2023mevis}, Ref-Youtube-VOS~\cite{seo2020urvos}, and Ref-DAVIS~\cite{khoreva2019video}. These results illustrate how much transferable video understanding each training dataset provides. The baseline model used in the comparisons is Sa2VA~\cite{yuan2025sa2va}.

Notably, the model trained on \dataset achieves the highest performance across all benchmarks, demonstrating the strong generalization capability of our interaction-centric data. Although these benchmarks primarily feature isolated object descriptions and insufficient interaction cues, \dataset still facilitates the learning of robust visual-language alignment.
We also report the results of our proposed architecture, \arch-1B, trained on the same \dataset data.
Although not explicitly designed solely for interaction-aware segmentation, \arch-1B effectively handles such cases while also performing competitively on standard RVOS benchmarks, demonstrating the generalizability of our framework.

\subsection{Qualitative results}
\label{suppl:qual}
\begin{figure*}[t]
    \centering
    \includegraphics[width=0.7\textwidth]{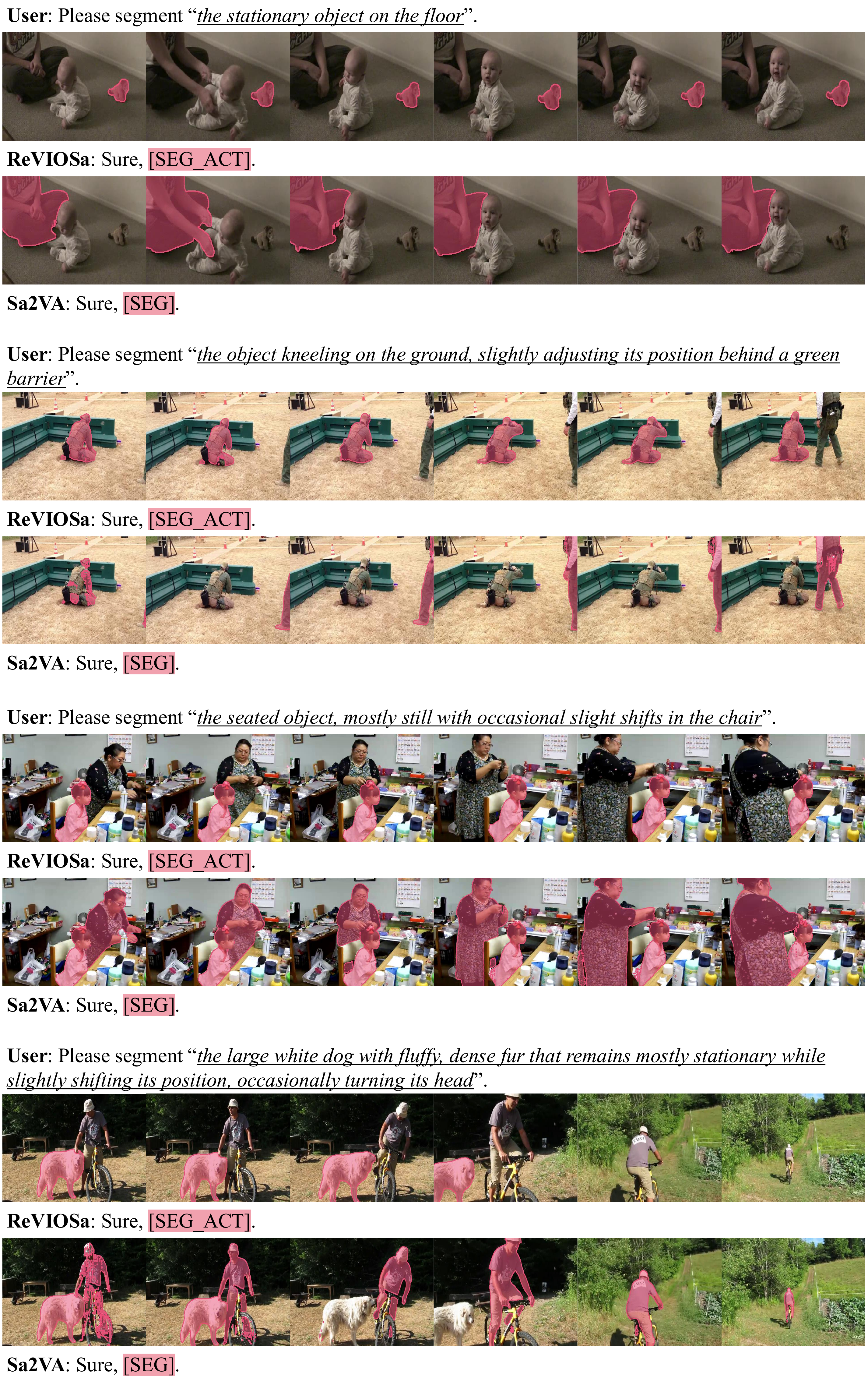}
    \caption{\textbf{Qualitative results.} Qualitative comparisons between our model (\arch) and the baseline model (Sa2VA) on the proposed \dataset dataset for the RVOS task. \ours consistently produces more accurate and temporally consistent segmentation masks, especially in challenging scenarios involving fine-grained interactions, appearance ambiguity, or motion. These results demonstrate the effectiveness of \ours in aligning linguistic cues with visual targets across time.}
    \label{fig:suppl_qual}
\end{figure*}

We present qualitative results to demonstrate the effectiveness of our proposed model in handling complex, interaction-centric referring expressions. Figure~\ref{fig:suppl_qual} compares our model (\arch) with a strong baseline (Sa2VA) on the proposed \dataset dataset for the RVOS task. Across a range of challenging scenarios involving ambiguous appearance, subtle motion, and fine-grained interactions, \arch consistently achieves more accurate and temporally consistent segmentation results. Notably, it exhibits strong alignment between the visual targets and the language expressions.

\begin{figure*}[t]
    \centering
    \includegraphics[width=0.7\textwidth]{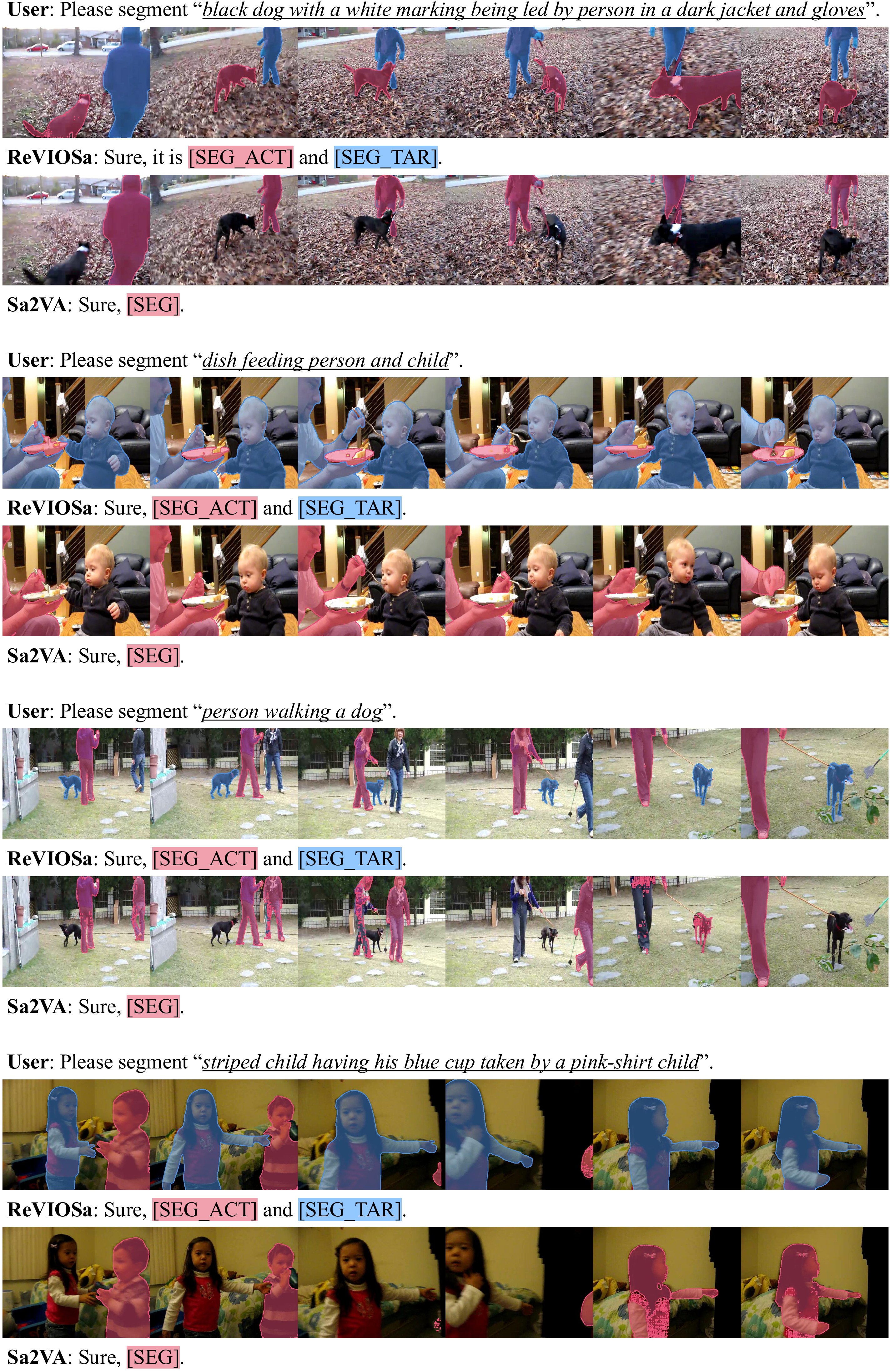}
    \caption{
    \textbf{Qualitative results.} Joint actor-target segmentation results using our proposed model with interaction-centric referring expressions on the \dataset dataset. Leveraging dedicated \actor and \target tokens, our model is able to segment both the actor (\qualpink{pink}) and the target (\qualblue{blue}) entities within a single forward pass. Each example corresponds to a complex expression describing an interaction between two entities. These results demonstrate the model’s ability to localize and distinguish multiple semantically linked objects simultaneously, showing potential for downstream applications such as human-object interaction understanding, social activity recognition, and fine-grained video scene interpretation.}
    \label{fig:suppl_app0}
\end{figure*}

\begin{figure*}[t]
    \centering
    \includegraphics[width=0.7\textwidth]{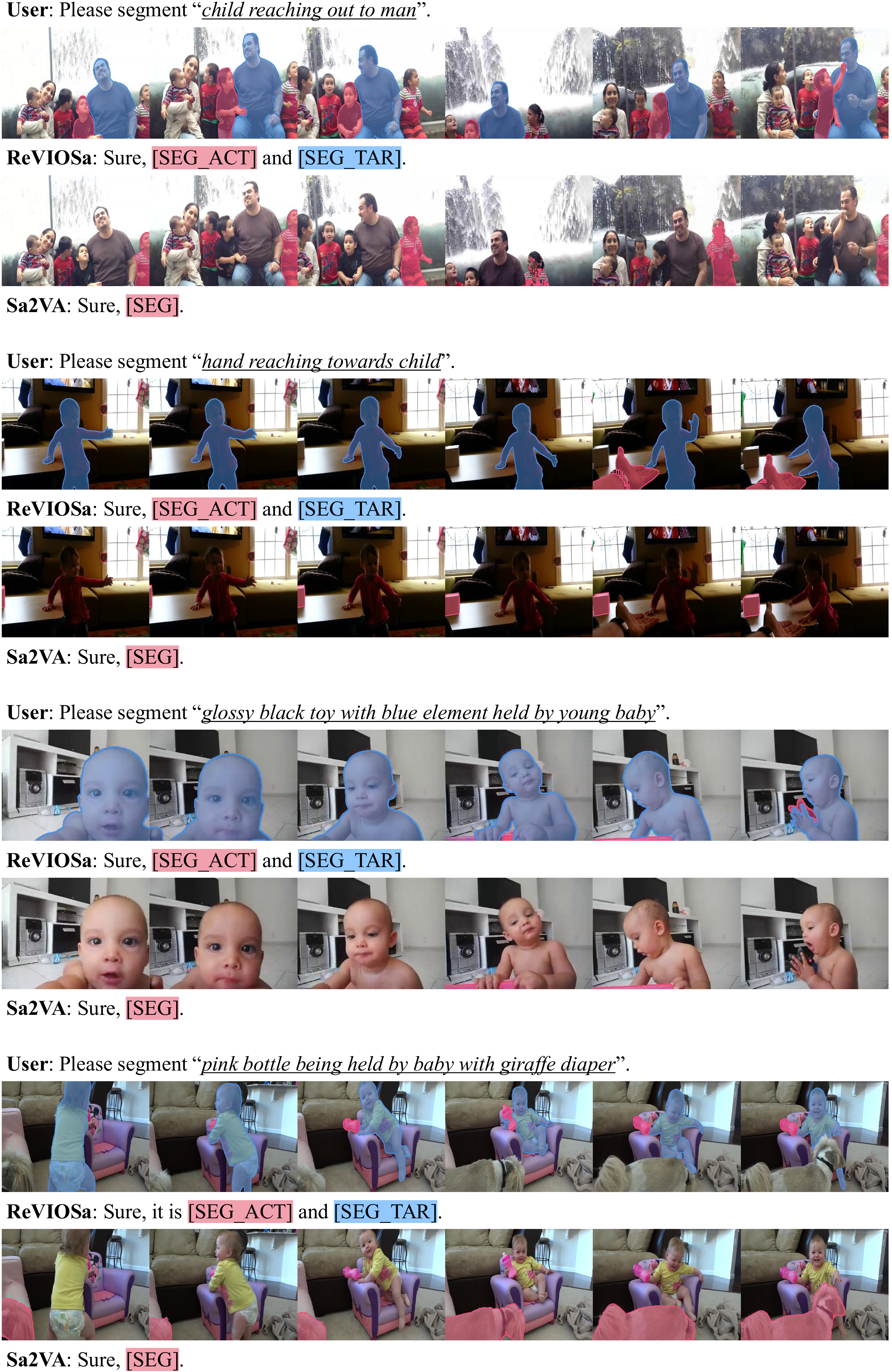}
    \caption{
    \textbf{Qualitative results.} Joint actor-target segmentation results using our proposed model with interaction-centric referring expressions on the \dataset dataset. Leveraging dedicated \actor and \target tokens, our model is able to segment both the actor (\qualpink{pink}) and the target (\qualblue{blue}) entities within a single forward pass. Each example corresponds to a complex expression describing an interaction between two entities. These results demonstrate the model’s ability to localize and distinguish multiple semantically linked objects simultaneously, showing potential for downstream applications such as human-object interaction understanding, social activity recognition, and fine-grained video scene interpretation.}
    \label{fig:suppl_app1}
\end{figure*}

In addition to standard referring segmentation, our model is also designed to perform joint subject-object inference within a single forward pass. As illustrated in Figure~\ref{fig:suppl_app0} and~\ref{fig:suppl_app1}, the model utilizes dedicated \actor and \target tokens to simultaneously localize both the subject and the object described in interaction-centric expressions. This dual segmentation capability enables our model to effectively capture relational semantics and dynamic interactions between entities. Such ability opens up opportunities for downstream applications such as human-object interaction understanding, social behavior analysis, and fine-grained activity reasoning in videos.

These qualitative results collectively validate the robustness, flexibility, and extensibility of our approach in real-world video understanding tasks that require precise multi-entity segmenting guided by natural language.

\section{Additional details of \dataset}
\label{suppl:dataset_detail}

\subsection{Data annotation pipeline}
\label{suppl:annotation_pipeline}

Our automatic data annotation pipeline consist of four-stage process. Among these, \textbf{Stage 1} and \textbf{Stage 3} utilize GPT-4o~\cite{hurst2024gpt} to extract accurate object-level and interaction-level information from video contexts. In contrast, \textbf{Stage 2} and \textbf{Stage 4} focus on converting this structured information into natural language referring expressions, for which we employ the quantized version of the LLaMA 3.1 Instruct model~\cite{grattafiori2024llama}.

To complement the overview in the main paper, we provide a more detailed explanation of the annotation pipeline here. Our stage-wise design enables a progressive buildup of annotation complexity, from basic object-level descriptions to more complex interaction-aware expressions.

\paragraph{Stage 1: Single object information.} In the first stage, we focus on individual objects to obtain rich descriptions encompassing both appearance and motion attributes. We highlight a single object within the video frame and give as an input, then GPT generates comprehensive object-centric captions that form the foundation for downstream stages. These descriptions ensure that each object is sufficiently characterized before reasoning about their interactions.

\paragraph{Stage 2: Single and multi-instance referring expressions.} In this stage, the captions obtained from Stage 1 are reformulated into referring expressions. We handle both single object and multi-instance cases:
(1) Single object expressions are generated by separating the original caption into three distinct types: appearance-only, motion-only, and combined (appearance and motion), offering finer-grained reference diversity.
(2) Multi-instance expressions are created by analyzing motion similarities across objects. If multiple objects exhibit similar motion patterns, the model decide whether to merge them into a single referring expression, thereby supporting both atomic and collective object references.

\paragraph{Stage 3: Interaction information.} In the third stage, we explore potential interactions among multiple objects within the video. Each object is annotated with an index label (e.g., \texttt{[0]}, \texttt{[1]}) and fed into GPT to assess whether interactions are present. If interactions exist, we further distinguish between two types:
(1) Unidirectional interactions, where a clear actor-target relationship exists (e.g., “Object \texttt{[0]} is leaning against object \texttt{[2]}"). For each pair, we generate two pseudo-captions with roles reversed (e.g., “Object \texttt{[2]} is being leaned on by object \texttt{[0]}") and extract structured actor-target mappings.
(2) Bidirectional interactions, where objects participate equally (e.g., “Object \texttt{[0]} and object \texttt{[1]} are standing together with arms around each other"). In such cases, only the object pair involved is extracted without role assignment.
This stage is crucial for capturing the relational structure between entities and building a pool of interaction data that reflects both directionality and symmetry.

\begin{table*}[!t]
    \centering
    \begin{tabular}{l|cccccc}
    \toprule
    \multirow{2}{*}{}
    \multirow{2}{*}{Datasets} & \multirow{2}{*}{Video} & \multirow{2}{*}{Object} & \multirow{2}{*}{Expression} & \multirow{2}{*}{Object/Video} & Actor-Target \\
     &  &  &  &  & Interaction \\
    \midrule
    A2D Sentence~\citep{gavrilyuk2018actor} & 3,782 & 4,825 & 6,656 & 1.28 & - \\
    J-HMDB Sentence~\citep{gavrilyuk2018actor} & 928 & 928 & 928 & 1 & - \\
    Ref-DAVIS~\citep{khoreva2019video} & 90 & 205 & 1,544 & 2.27 & - \\
    Ref-Youtube-VOS~\citep{seo2020urvos} & 3,978 & 7,451 & 15,009 & 1.86 & - \\
    MeViS~\citep{ding2023mevis} & 2,006 & 8,171 & 28,570 & \underline{4.28} & - \\
    ReVOS~\citep{yan2024visa} & 1,042 & 5,535 & 35,074 & 5.31 & - \\
    Ref-SAV~\citep{yuan2025sa2va} & 37,311 & 72,509 & 72,509 & 1.94 & - \\
    \midrule
    \textbf{\dataset (Ours)} & 8,738 & 35,247 & 127,236 & 4.03 & 17,604 \\
    \bottomrule
    \end{tabular}
    \caption{\textbf{Comparison of various RVOS datasets.} Our newly proposed \textbf{\dataset} offers the largest number of referring expressions and a high object-per-video ratio, enabling richer and more diverse visual grounding across complex scenes compared to existing benchmarks. Unlike existing datasets, \dataset also provides interaction-aware referring expressions that explicitly distinguish between actor and target roles, enabling fine-grained understanding of visual interactions.}
    \label{tab:suppl_statistics}
\end{table*}

\paragraph{Stage 4: Interaction-aware referring expressions.} In the final stage, we convert structured interaction information from Stage 3 into rich referring expressions. Starting from GPT-generated index-based captions (e.g., “Object \texttt{[0]} is leaning against object \texttt{[2]}"), we inject class and appearance description for each object obtained from stage 2 to produce semantically enriched expressions. This yields two levels of interaction captions:
(1) Class-level, using coarse object category labels
(2) Appearance-level, incorporating visual attributes from earlier stages.

Throughout the entire data annotation pipeline, the \dataset dataset evolves into a diverse and large-scale resource that simultaneously provides rich descriptions of object interactions, ranging from simple to highly detailed expressions.

\subsection{Additional examples of \dataset}
\label{suppl:data_samples}

\begin{figure*}[t]
    \centering
    \includegraphics[width=0.9\textwidth]{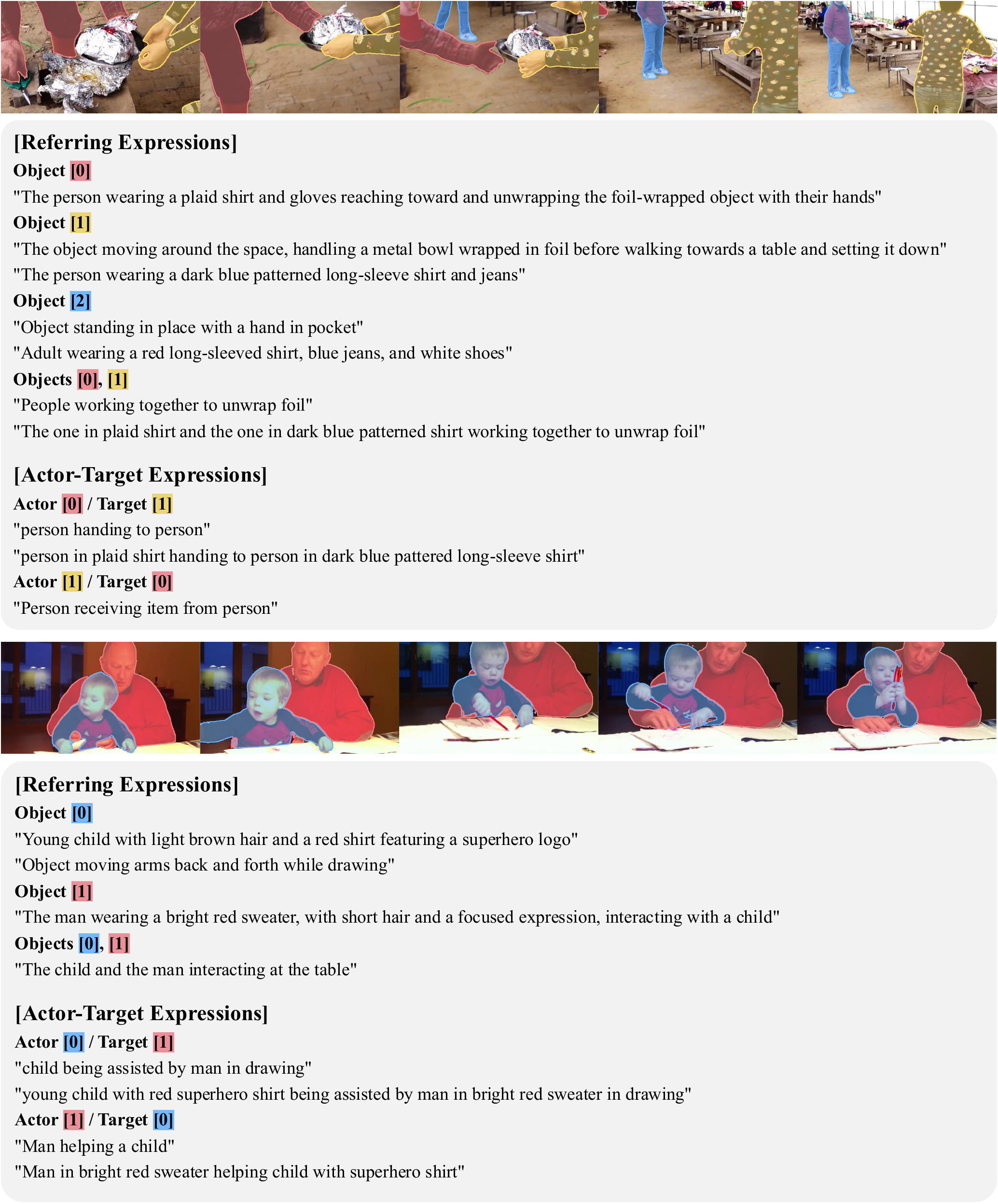}
    \caption{\textbf{Examples of \dataset.}}
    \label{fig:suppl_data_samples0}
\end{figure*}

\begin{figure*}[t]
    \centering
    \includegraphics[width=0.9\textwidth]{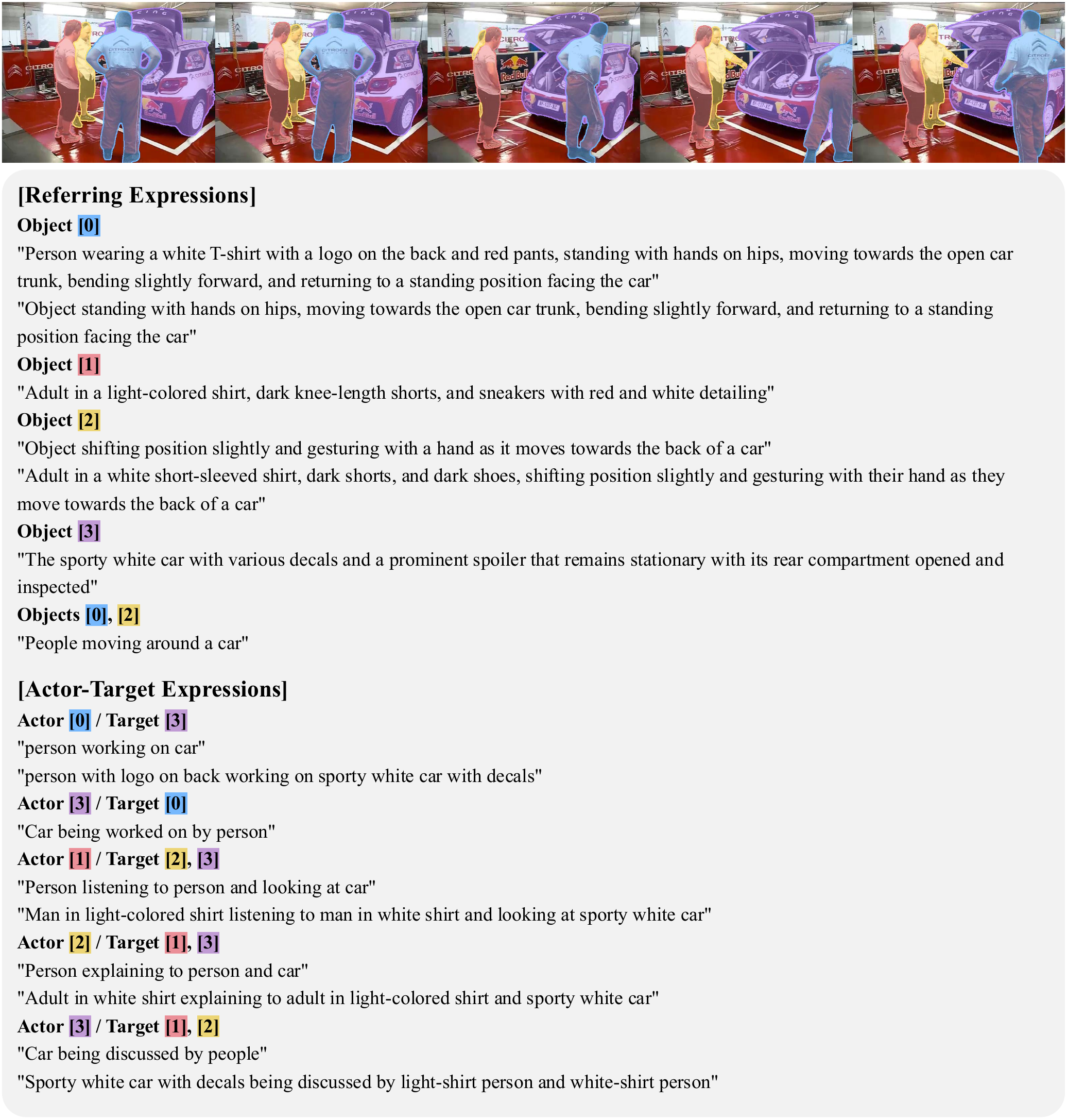}
    \caption{\textbf{Examples of \dataset.}}
    \label{fig:suppl_data_samples1}
\end{figure*}

Figure~\ref{fig:suppl_data_samples0} and Figure~\ref{fig:suppl_data_samples1} present additional examples from the \dataset dataset. Our dataset covers a broad range of referring expressions, including challenging cases like multi-object references and motion-only descriptions, as well as varying levels of granularity from class-level to fine-grained appearance-based expressions.
It also includes interaction-focused expressions that clearly distinguish actor and target roles. The examples illustrate multiple objects within a single video and their relationships, highlighting the dataset’s ability to capture object-level interactions in complex scenes.

\subsection{Video clip extraction procedure}
\label{suppl:video_clip}
The \dataset dataset is constructed using source videos from the VidOR dataset~\cite{shang2019annotating}, which contains a large number of long-form videos, many exceeding 1,000 frames in length. To generate more diverse and effective video clips for referring video object segmentation, we apply a systematic clip extraction strategy. Specifically, each original source video is divided into non-overlapping temporal bins of 1,000 frames. From these, we select only the first and last bins to increase the likelihood of capturing distinct scenes or transitions within a single video. Within each selected bin, we extract only the first 500 frames to form a video clip. This approach allows us to generate a wide range of video segments while ensuring sufficient temporal context and diverse scene required for RVOS. As a result, we obtain high-quality video clips that are both temporally coherent and suitable for dense language grounding and interaction modeling.

\subsection{Dataset statistics}
\label{suppl:statistics}
\begin{figure*}[t]
    \centering
    \includegraphics[width=0.75\textwidth]{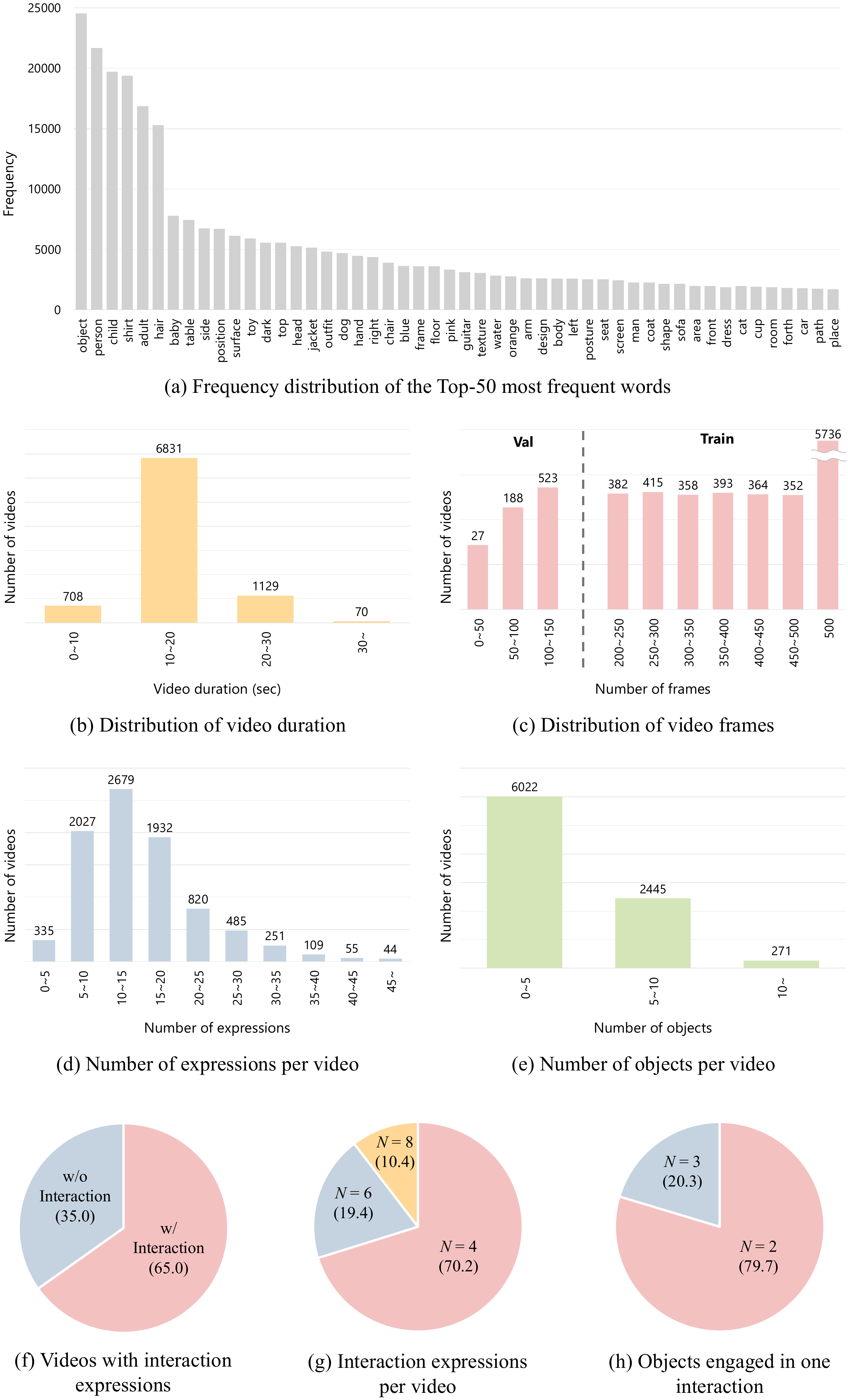}
    \caption{\textbf{Overall statistics of \dataset.}}
    \label{fig:suppl_data_statistics}
\end{figure*}

The overall statistics of the \dataset dataset are presented in Figure~\ref{fig:suppl_data_statistics}, with a brief comparison of statistics across datasets provided in Table~\ref{tab:suppl_statistics}.
The word frequency distribution (a) reveals that commonly used terms such as \textit{object}, \textit{person}, \textit{child}, \textit{side}, \textit{position}, and \textit{right} frequently appear in the referring expressions. This indicates that the dataset captures not only static appearance information but also emphasizes spatial relations and interactive contexts involving everyday entities. In terms of temporal characteristics, (b) shows that most videos fall within the 10 to 20 second range, providing sufficient temporal context for modeling object-level dynamics. Additionally, (c) illustrates the distribution of video frames: the training set mostly consists of 500 frames, while the validation set is composed of shorter clips with frame counts aligned in increments of 5.

The dataset also exhibits significant linguistic density and visual complexity. As shown in (d), most videos are annotated with 5 to 20 referring expressions, peaking at the 10 to 15 range, which enables dense language grounding for each clip. Moreover, (e) indicates that a large portion of videos contain 0 to 5 annotated objects, with a smaller but meaningful subset containing more than 5. This diversity in object count allows the dataset to cover a broad range of scene complexities, from simple to highly interactive scenarios. Collectively, these statistics confirm that the \dataset dataset is well-suited for advancing research in referring video object segmentation and interaction-centric video understanding.

Furthermore, (f), (g), (h) provides an overall interaction-focused statistics within \dataset. In (f), we observe that approximately 65\% of videos contain at least one interaction-based referring expression, indicating that interaction scenarios are prevalent throughout the dataset. (g) further illustrates the distribution of the number of interaction expressions per video, and (h) shows the number of objects involved in each interaction; while most interactions involve two objects, a notable 20.3\% involve three, suggesting a considerable portion of the dataset covers more complex, multi-object interactions.

\begin{figure*}[t]
    \centering
    \includegraphics[width=0.8\textwidth]{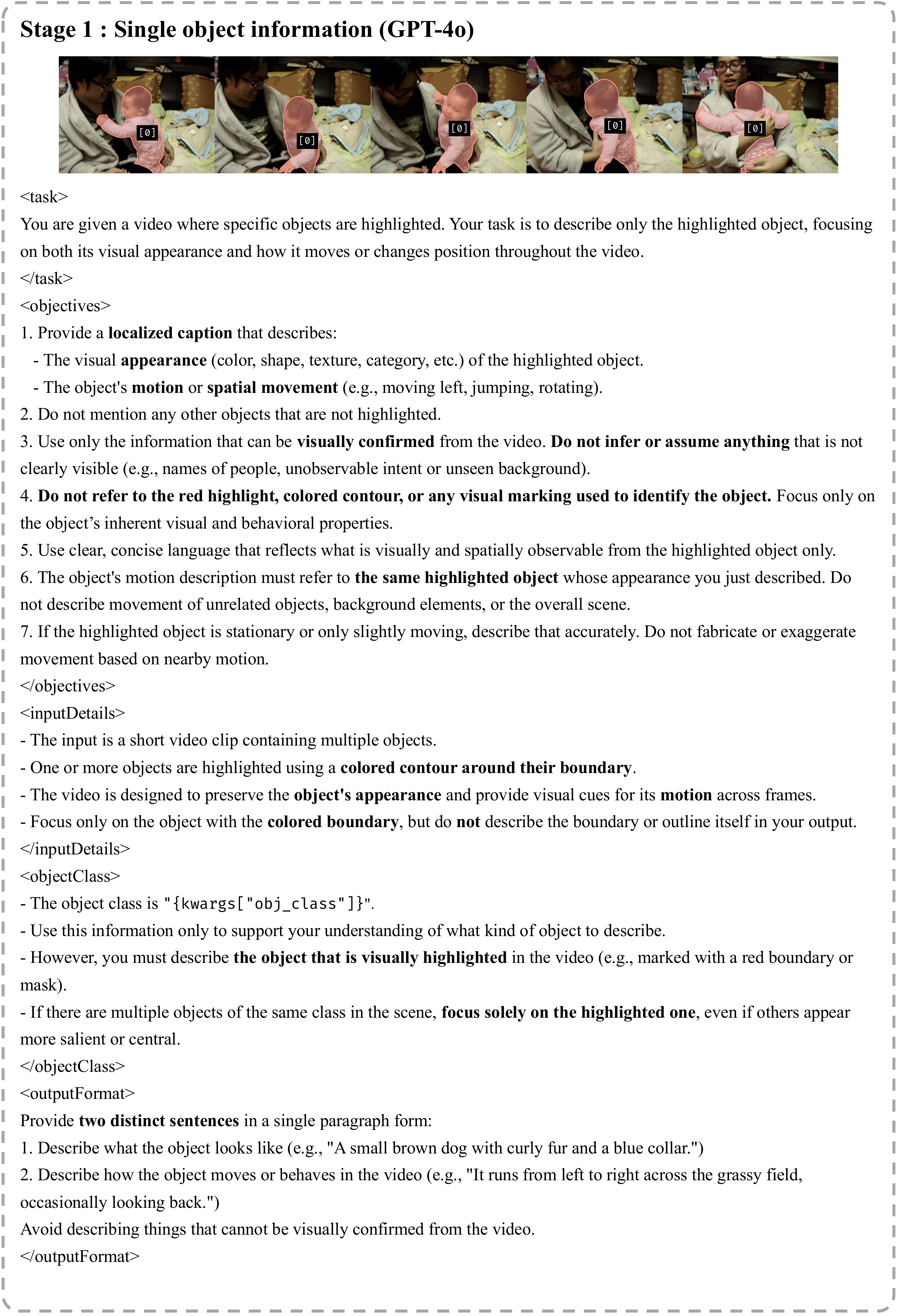}
    \caption{\textbf{Stage 1: Input prompts to GPT-4o.} We provide GPT-4o with preprocessed video frames in which objects are highlighted using labels and colored masks. This stage aims to extract localized information for each object, including both appearance and motion attributes.}
    \label{fig:suppl_prompt_1}
\end{figure*}

\begin{figure*}[t]
    \centering
    \includegraphics[width=0.8\textwidth]{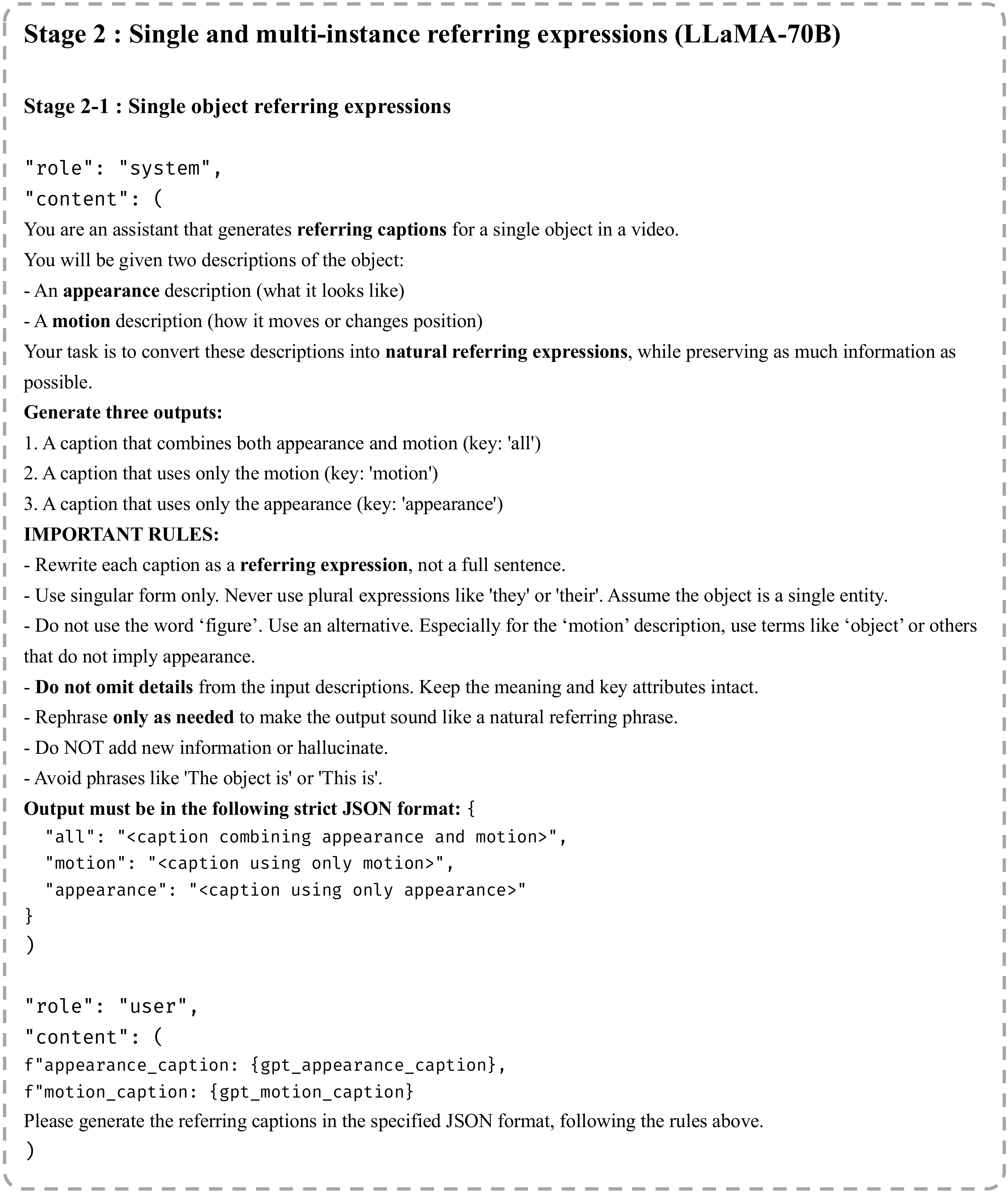}
    \caption{\textbf{Stage 2 (Single-object case): Input prompts to LLaMA.} Using the object-level descriptions generated in Stage 1, we prompt LLaMA to produce diverse referring expressions. For single-object cases, we decompose the description into three types: appearance-only, motion-only, and combined expressions.}
    \label{fig:suppl_prompt_2}
\end{figure*}

\begin{figure*}[t]
    \centering
    \includegraphics[width=0.8\textwidth]{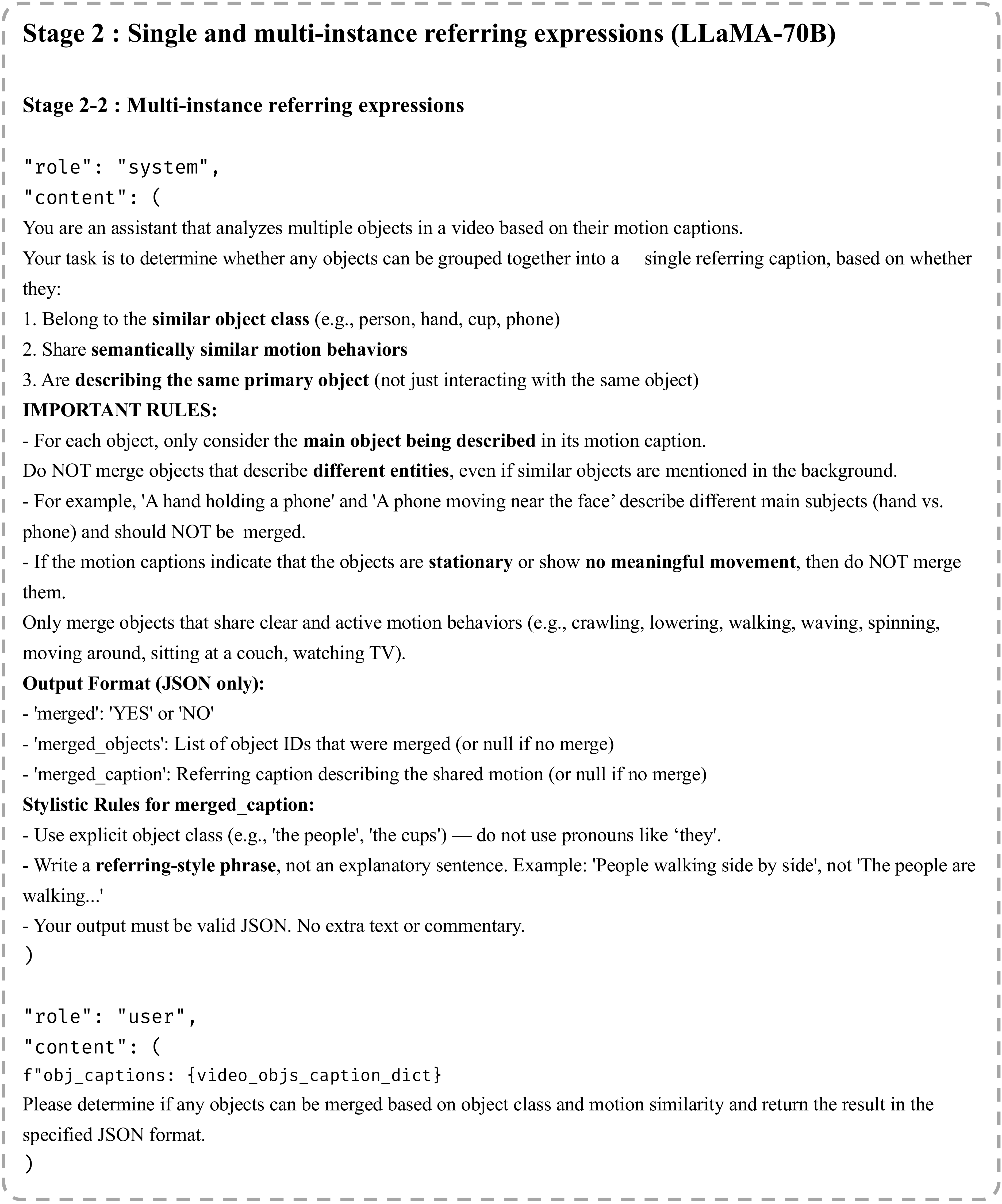}
    \caption{\textbf{Stage 2 (Multi-instance case): Input prompts to LLaMA.} For videos containing multiple objects with similar motion, we prompt LLaMA to determine whether they should be merged into a single referring expression. The decision is made based on motion similarity.}
    \label{fig:suppl_prompt_3}
\end{figure*}

\begin{figure*}[t]
    \centering
    \includegraphics[width=0.8\textwidth]{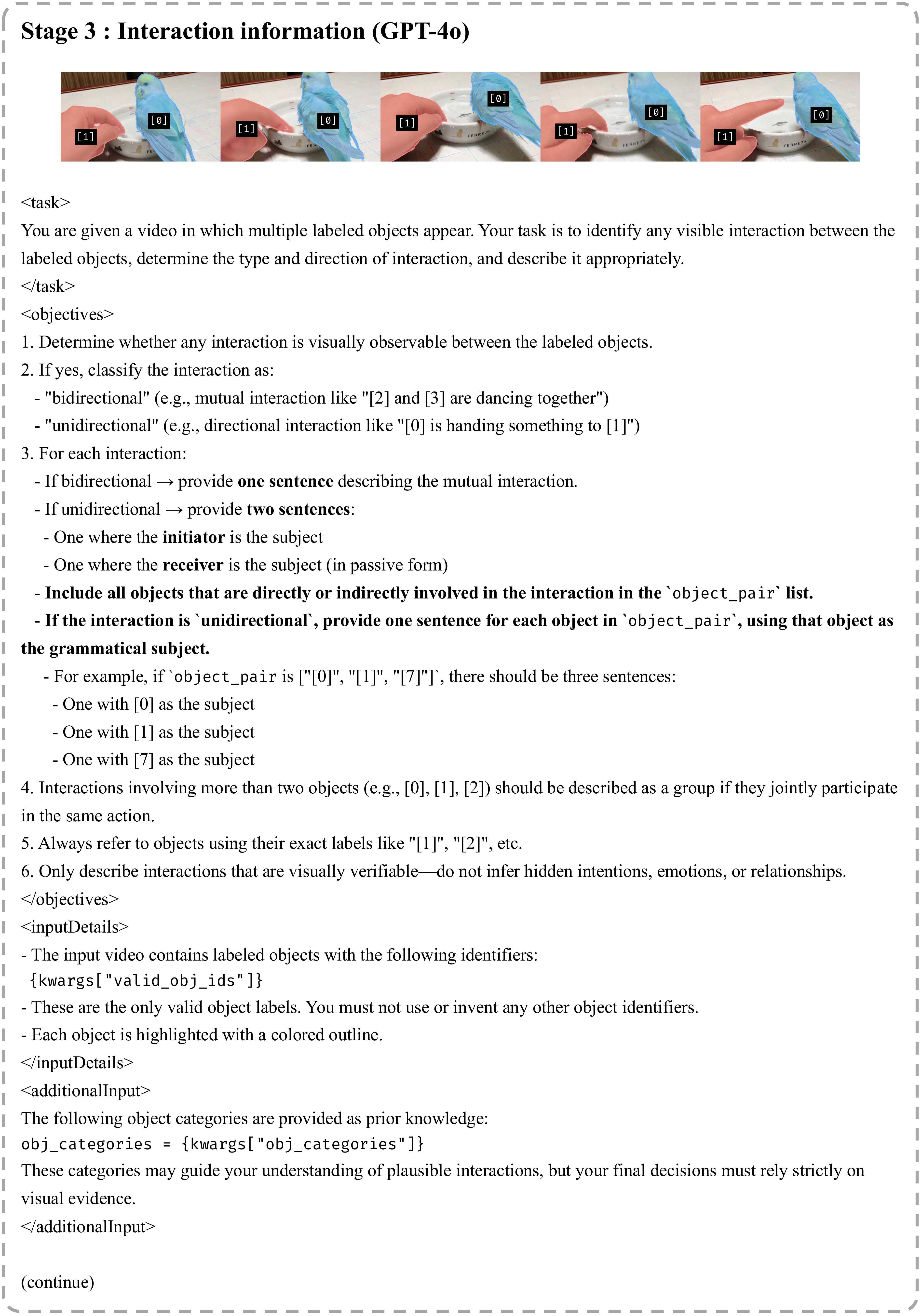}
    \caption{\textbf{Stage 3: Input prompts to GPT-4o.} We provide GPT-4o with preprocessed frames highlighting all objects with labels and colored masks. This stage focuses on detecting interactions between objects and generating detailed descriptions of their relationships.}
    \label{fig:suppl_prompt_4}
\end{figure*}

\begin{figure*}[t]
    \centering
    \includegraphics[width=0.8\textwidth]{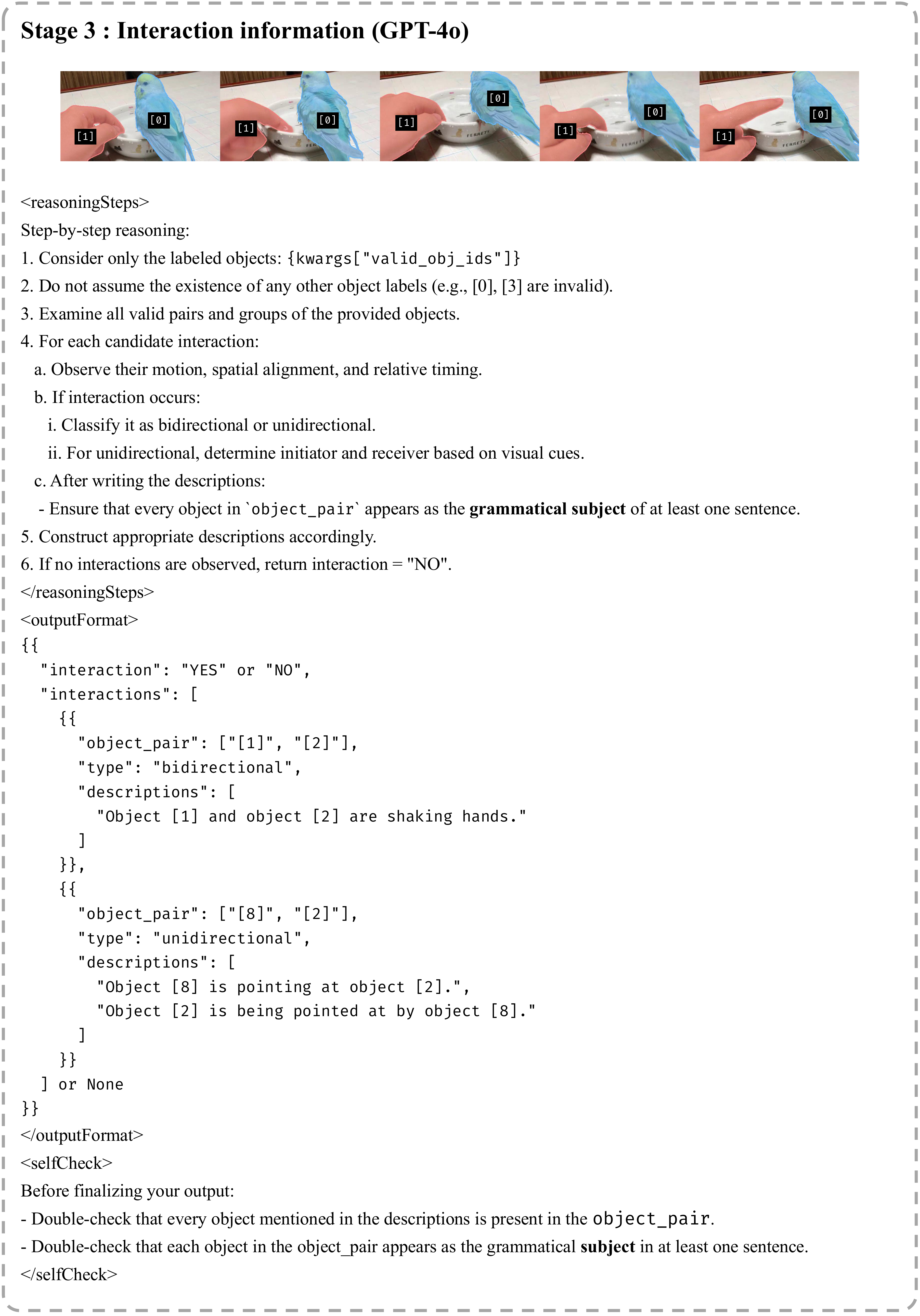}
    \caption{\textbf{Stage 3 : Input prompts to GPT-4o.} We provide GPT-4o with preprocessed frames highlighting all objects with labels and colored masks. This stage focuses on detecting interactions between objects and generating detailed descriptions of their relationships.}
    \label{fig:suppl_prompt_5}
\end{figure*}

\begin{figure*}[t]
    \centering
    \includegraphics[width=0.8\textwidth]{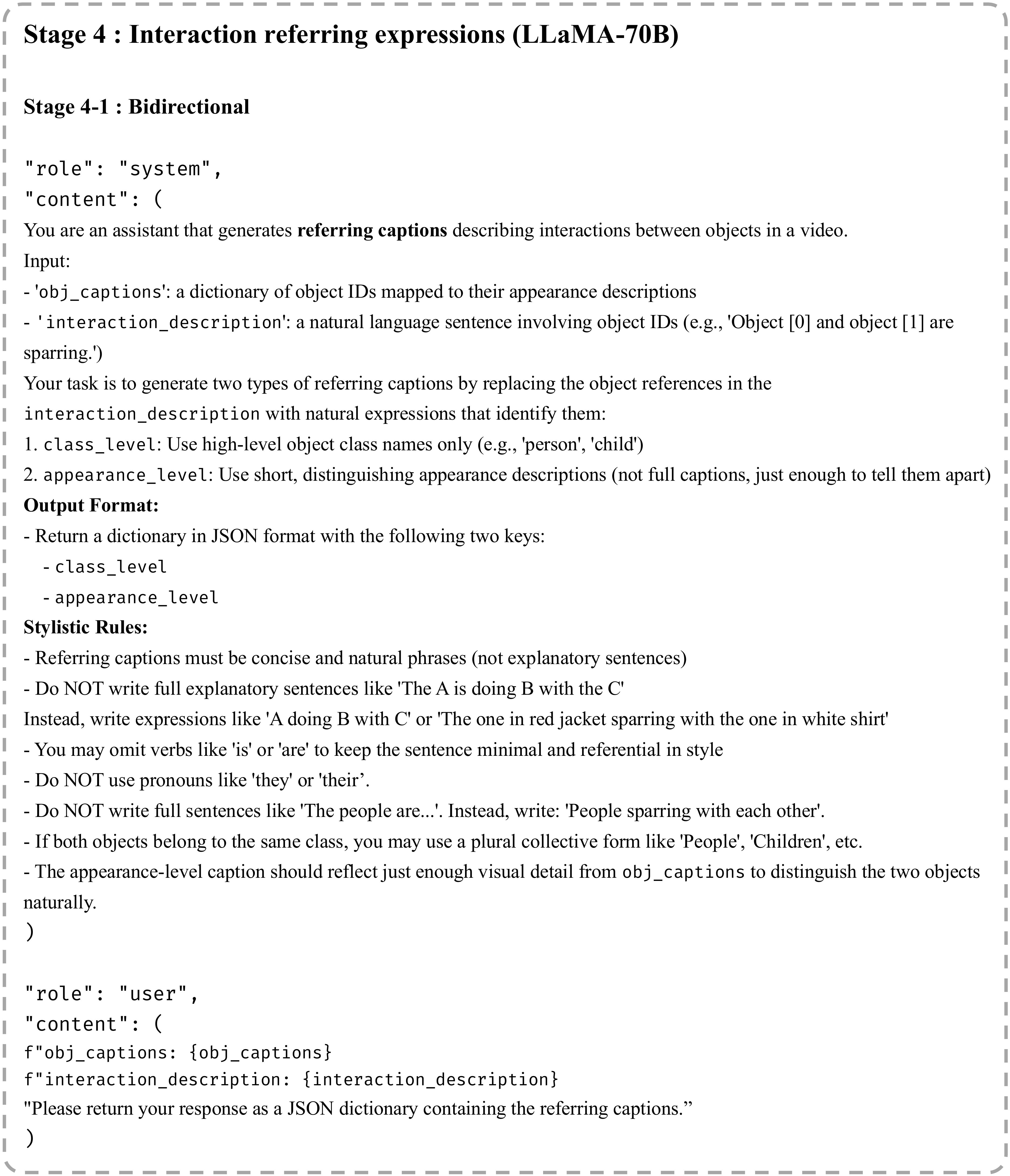}
    \caption{\textbf{Stage 4 (Bidirectional case): Input prompts to LLaMA.} We prompt LLaMA using interaction-level descriptions generated in Stage 3. Appearance and class information from Stage 2 are injected into each entity, indicated by labeled placeholders (e.g., \texttt{[0]}).}
    \label{fig:suppl_prompt_6}
\end{figure*}

\begin{figure*}[t]
    \centering
    \includegraphics[width=0.8\textwidth]{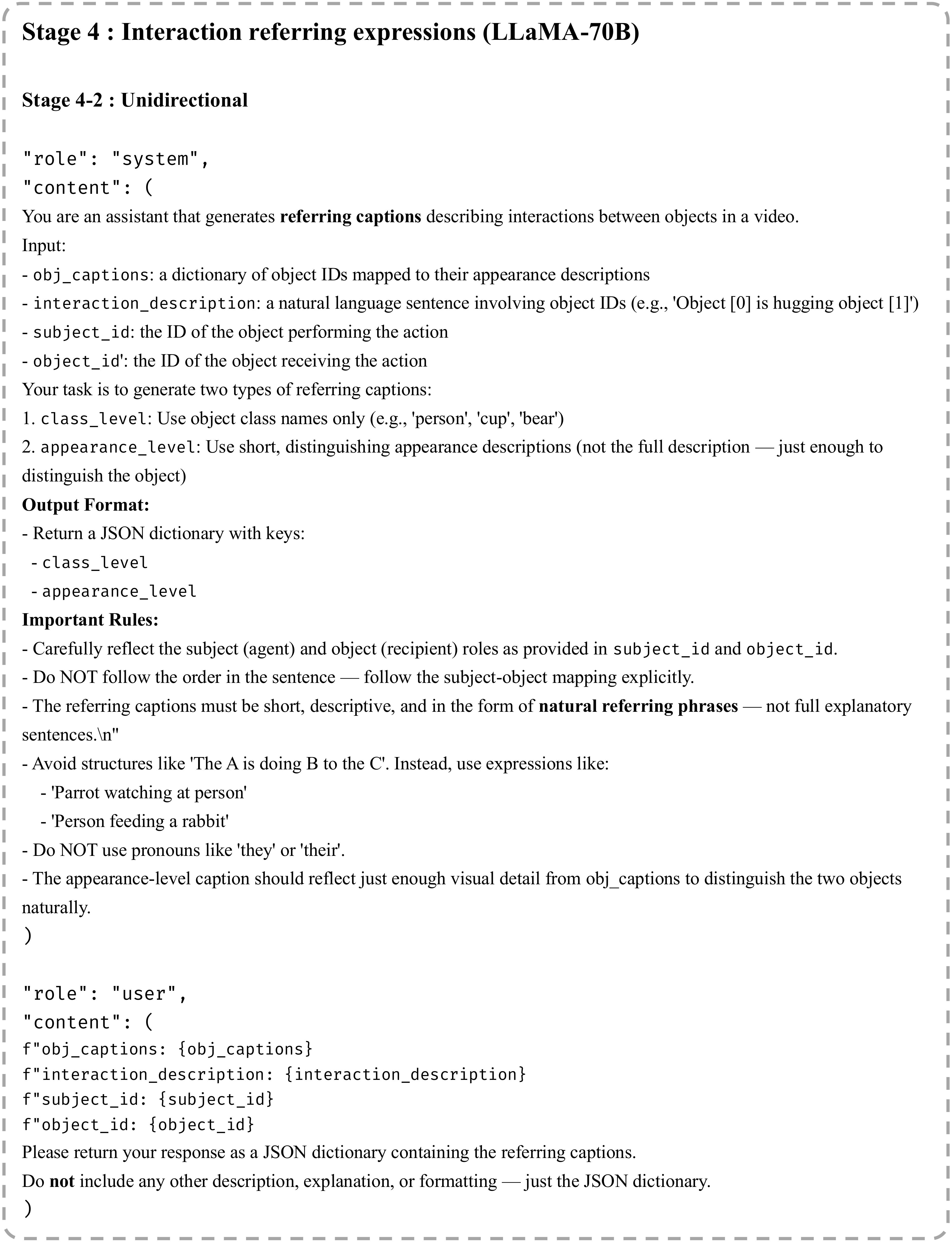}
    \caption{\textbf{Stage 4 (Unidirectional case): Input prompts to LLaMA.} In cases where the interaction is classified as \textit{unidirectional}, LLaMA additionally predicts actor object and target object identifiers. This enables us to assign distinct segmentation mask tracks to each role.}
    \label{fig:suppl_prompt_7}
\end{figure*}

\FloatBarrier

\bibliography{aaai2026}

\begin{thebibliography}{33}
\providecommand{\natexlab}[1]{#1}

\bibitem[{Bai et~al.(2024)Bai, He, Mei, Wang, Gao, Chen, Zhang, and Shou}]{bai2024one}
Bai, Z.; He, T.; Mei, H.; Wang, P.; Gao, Z.; Chen, J.; Zhang, Z.; and Shou, M.~Z. 2024.
\newblock One token to seg them all: Language instructed reasoning segmentation in videos.
\newblock \emph{Advances in Neural Information Processing Systems}, 37: 6833--6859.

\bibitem[{Botach, Zheltonozhskii, and Baskin(2022)}]{botach2022end}
Botach, A.; Zheltonozhskii, E.; and Baskin, C. 2022.
\newblock End-to-end referring video object segmentation with multimodal transformers.
\newblock In \emph{Proceedings of the IEEE/CVF Conference on Computer Vision and Pattern Recognition}, 4985--4995.

\bibitem[{Chen et~al.(2024)Chen, Wang, Cao, Liu, Gao, Cui, Zhu, Ye, Tian, Liu et~al.}]{chen2024expanding}
Chen, Z.; Wang, W.; Cao, Y.; Liu, Y.; Gao, Z.; Cui, E.; Zhu, J.; Ye, S.; Tian, H.; Liu, Z.; et~al. 2024.
\newblock Expanding performance boundaries of open-source multimodal models with model, data, and test-time scaling.
\newblock \emph{arXiv preprint arXiv:2412.05271}.

\bibitem[{Cheng et~al.(2022)Cheng, Misra, Schwing, Kirillov, and Girdhar}]{cheng2022masked}
Cheng, B.; Misra, I.; Schwing, A.~G.; Kirillov, A.; and Girdhar, R. 2022.
\newblock Masked-attention mask transformer for universal image segmentation.
\newblock In \emph{Proceedings of the IEEE/CVF conference on computer vision and pattern recognition}, 1290--1299.

\bibitem[{Ding et~al.(2023)Ding, Liu, He, Jiang, and Loy}]{ding2023mevis}
Ding, H.; Liu, C.; He, S.; Jiang, X.; and Loy, C.~C. 2023.
\newblock MeViS: A large-scale benchmark for video segmentation with motion expressions.
\newblock In \emph{Proceedings of the IEEE/CVF international conference on computer vision}, 2694--2703.

\bibitem[{Ding et~al.(2021)Ding, Liu, Wang, and Jiang}]{vision-language-transformer}
Ding, H.; Liu, C.; Wang, S.; and Jiang, X. 2021.
\newblock Vision-Language Transformer and Query Generation for Referring Segmentation.
\newblock In \emph{Proceedings of the IEEE International Conference on Computer Vision}.

\bibitem[{Fan et~al.(2021)Fan, Liu, Wang, Xu, Lin, Yuille, and Loy}]{fan2021moma}
Fan, Q.; Liu, Y.; Wang, W.; Xu, N.; Lin, D.; Yuille, A.; and Loy, C.~C. 2021.
\newblock MOMA: A Multi-Object Multi-Action Dataset for Understanding Human Activities.
\newblock In \emph{Advances in Neural Information Processing Systems (NeurIPS)}.

\bibitem[{Gavrilyuk et~al.(2018)Gavrilyuk, Ghodrati, Li, and Snoek}]{gavrilyuk2018actor}
Gavrilyuk, K.; Ghodrati, A.; Li, Z.; and Snoek, C.~G. 2018.
\newblock Actor and action video segmentation from a sentence.
\newblock In \emph{Proceedings of the IEEE conference on computer vision and pattern recognition}, 5958--5966.

\bibitem[{Grattafiori et~al.(2024)Grattafiori, Dubey, Jauhri, Pandey, Kadian, Al-Dahle, Letman, Mathur, Schelten, Vaughan et~al.}]{grattafiori2024llama}
Grattafiori, A.; Dubey, A.; Jauhri, A.; Pandey, A.; Kadian, A.; Al-Dahle, A.; Letman, A.; Mathur, A.; Schelten, A.; Vaughan, A.; et~al. 2024.
\newblock The llama 3 herd of models.
\newblock \emph{arXiv e-prints}, arXiv--2407.

\bibitem[{Hu et~al.(2022)Hu, Shen, Wallis, Allen-Zhu, Li, Wang, Wang, Chen et~al.}]{hu2022lora}
Hu, E.~J.; Shen, Y.; Wallis, P.; Allen-Zhu, Z.; Li, Y.; Wang, S.; Wang, L.; Chen, W.; et~al. 2022.
\newblock Lora: Low-rank adaptation of large language models.
\newblock \emph{ICLR}, 1(2): 3.

\bibitem[{Hurst et~al.(2024)Hurst, Lerer, Goucher, Perelman, Ramesh, Clark, Ostrow, Welihinda, Hayes, Radford et~al.}]{hurst2024gpt}
Hurst, A.; Lerer, A.; Goucher, A.~P.; Perelman, A.; Ramesh, A.; Clark, A.; Ostrow, A.; Welihinda, A.; Hayes, A.; Radford, A.; et~al. 2024.
\newblock Gpt-4o system card.
\newblock \emph{arXiv preprint arXiv:2410.21276}.

\bibitem[{Ji et~al.(2020)Ji, Wang, Xu, Qi, Zhu, and Zhu}]{ji2020action}
Ji, M.; Wang, J.; Xu, X.; Qi, S.; Zhu, Y.; and Zhu, S.-C. 2020.
\newblock Action Genome: Actions as Compositions of Spatiotemporal Scene Graphs.
\newblock In \emph{IEEE Conference on Computer Vision and Pattern Recognition (CVPR)}, 10236--10247.

\bibitem[{Khoreva, Rohrbach, and Schiele(2019)}]{khoreva2019video}
Khoreva, A.; Rohrbach, A.; and Schiele, B. 2019.
\newblock Video object segmentation with language referring expressions.
\newblock In \emph{Computer Vision--ACCV 2018: 14th Asian Conference on Computer Vision, Perth, Australia, December 2--6, 2018, Revised Selected Papers, Part IV 14}, 123--141. Springer.

\bibitem[{Lai et~al.(2024)Lai, Tian, Chen, Li, Yuan, Liu, and Jia}]{lai2024lisa}
Lai, X.; Tian, Z.; Chen, Y.; Li, Y.; Yuan, Y.; Liu, S.; and Jia, J. 2024.
\newblock Lisa: Reasoning segmentation via large language model.
\newblock In \emph{Proceedings of the IEEE/CVF Conference on Computer Vision and Pattern Recognition}, 9579--9589.

\bibitem[{Liang et~al.(2021)Liang, Wu, Zhou, Wang, Yang, Wei, and Yang}]{liang2021rethinking}
Liang, C.; Wu, Y.; Zhou, T.; Wang, W.; Yang, Z.; Wei, Y.; and Yang, Y. 2021.
\newblock Rethinking cross-modal interaction from a top-down perspective for referring video object segmentation.
\newblock \emph{arXiv preprint arXiv:2106.01061}.

\bibitem[{Liu et~al.(2023)Liu, Li, Wu, and Lee}]{liu2023visual}
Liu, H.; Li, C.; Wu, Q.; and Lee, Y.~J. 2023.
\newblock Visual instruction tuning.
\newblock \emph{Advances in neural information processing systems}, 36: 34892--34916.

\bibitem[{Miao et~al.(2023)Miao, Bennamoun, Gao, and Mian}]{miao2023spectrum}
Miao, B.; Bennamoun, M.; Gao, Y.; and Mian, A. 2023.
\newblock Spectrum-guided multi-granularity referring video object segmentation.
\newblock In \emph{Proceedings of the IEEE/CVF International Conference on Computer Vision}, 920--930.

\bibitem[{Munasinghe et~al.(2025)Munasinghe, Gani, Zhu, Cao, Xing, Khan, and Khan}]{munasinghe2025videoglamm}
Munasinghe, S.; Gani, H.; Zhu, W.; Cao, J.; Xing, E.; Khan, F.~S.; and Khan, S. 2025.
\newblock Videoglamm: A large multimodal model for pixel-level visual grounding in videos.
\newblock In \emph{Proceedings of the Computer Vision and Pattern Recognition Conference}, 19036--19046.

\bibitem[{Rasheed et~al.(2024)Rasheed, Maaz, Shaji, Shaker, Khan, Cholakkal, Anwer, Xing, Yang, and Khan}]{rasheed2024glamm}
Rasheed, H.; Maaz, M.; Shaji, S.; Shaker, A.; Khan, S.; Cholakkal, H.; Anwer, R.~M.; Xing, E.; Yang, M.-H.; and Khan, F.~S. 2024.
\newblock Glamm: Pixel grounding large multimodal model.
\newblock In \emph{Proceedings of the IEEE/CVF Conference on Computer Vision and Pattern Recognition}, 13009--13018.

\bibitem[{Ravi et~al.(2024)Ravi, Gabeur, Hu, Hu, Ryali, Ma, Khedr, R{\"a}dle, Rolland, Gustafson et~al.}]{ravi2024sam}
Ravi, N.; Gabeur, V.; Hu, Y.-T.; Hu, R.; Ryali, C.; Ma, T.; Khedr, H.; R{\"a}dle, R.; Rolland, C.; Gustafson, L.; et~al. 2024.
\newblock Sam 2: Segment anything in images and videos.
\newblock \emph{arXiv preprint arXiv:2408.00714}.

\bibitem[{Seo, Lee, and Han(2020)}]{seo2020urvos}
Seo, S.; Lee, J.-Y.; and Han, B. 2020.
\newblock Urvos: Unified referring video object segmentation network with a large-scale benchmark.
\newblock In \emph{Computer Vision--ECCV 2020: 16th European Conference, Glasgow, UK, August 23--28, 2020, Proceedings, Part XV 16}, 208--223. Springer.

\bibitem[{Shang et~al.(2017)Shang, Li, Zhang, Jiang, Yang, and Chen}]{shang2017video}
Shang, X.; Li, L.; Zhang, Y.; Jiang, T.; Yang, X.; and Chen, Z. 2017.
\newblock Video Relationship Detection.
\newblock In \emph{ACM International Conference on Multimedia (ACM MM)}, 1074--1082.

\bibitem[{Shang et~al.(2019)Shang, Ren, Li, Chen, Liu, and Zhou}]{shang2019annotating}
Shang, X.; Ren, X.; Li, L.; Chen, Z.; Liu, Y.-G.; and Zhou, Y. 2019.
\newblock Annotating Objects and Relations in User-Generated Videos.
\newblock In \emph{ACM International Conference on Multimedia (ACM MM)}, 1308--1316.

\bibitem[{Thomee et~al.(2016)Thomee, Shamma, Friedland, Elizalde, Ni, Poland, Borth, and Li}]{thomee2016yfcc100m}
Thomee, B.; Shamma, D.~A.; Friedland, G.; Elizalde, B.; Ni, K.; Poland, D.; Borth, D.; and Li, L.-J. 2016.
\newblock YFCC100M: The New Data in Multimedia Research.
\newblock \emph{Communications of the ACM}, 59(2): 64--73.

\bibitem[{Wang et~al.(2023)Wang, Gao, Zhu, and Dai}]{wang2023soc}
Wang, J.; Gao, F.; Zhu, J.; and Dai, Q. 2023.
\newblock SOC: Segmenting Objects by Categories for Referring Video Object Segmentation.
\newblock \emph{arXiv preprint arXiv:2305.17011}.

\bibitem[{Wang et~al.(2024)Wang, Ma, Li, and et~al.}]{wang2024villa}
Wang, R.; Ma, Z.; Li, X.; and et~al. 2024.
\newblock ViLLa: Unifying Vision-Language Segmentation Tasks with Large Multi-modal Models.
\newblock \emph{arXiv preprint arXiv:2407.14500}.

\bibitem[{Wu et~al.(2022{\natexlab{a}})Wu, Dong, Shao, and Shen}]{wu2022multi}
Wu, D.; Dong, X.; Shao, L.; and Shen, J. 2022{\natexlab{a}}.
\newblock Multi-level representation learning with semantic alignment for referring video object segmentation.
\newblock In \emph{Proceedings of the IEEE/CVF Conference on Computer Vision and Pattern Recognition}, 4996--5005.

\bibitem[{Wu et~al.(2022{\natexlab{b}})Wu, Jiang, Sun, Yuan, and Luo}]{wu2022language}
Wu, J.; Jiang, Y.; Sun, P.; Yuan, Z.; and Luo, P. 2022{\natexlab{b}}.
\newblock Language as Queries for Referring Video Object Segmentation.
\newblock In \emph{2022 IEEE/CVF Conference on Computer Vision and Pattern Recognition (CVPR)}, 4964--4974. IEEE.

\bibitem[{Wu et~al.(2024)Wu, Yu, Jia, Jin, Zhou, and Qiao}]{wu2024star}
Wu, Y.; Yu, G.; Jia, W.; Jin, Q.; Zhou, J.; and Qiao, Y. 2024.
\newblock STAR: Structured Action Understanding in Instructional Videos.
\newblock \emph{arXiv preprint arXiv:2405.09711}.

\bibitem[{Yan et~al.(2024)Yan, Wang, Yan, Jiang, Hu, Kang, Xie, and Gavves}]{yan2024visa}
Yan, C.; Wang, H.; Yan, S.; Jiang, X.; Hu, Y.; Kang, G.; Xie, W.; and Gavves, E. 2024.
\newblock Visa: Reasoning video object segmentation via large language models.
\newblock In \emph{European Conference on Computer Vision}, 98--115. Springer.

\bibitem[{Yuan et~al.(2025)Yuan, Li, Zhang, Huang, Xu, Ji, Tong, Qi, Feng, and Yang}]{yuan2025sa2va}
Yuan, H.; Li, X.; Zhang, T.; Huang, Z.; Xu, S.; Ji, S.; Tong, Y.; Qi, L.; Feng, J.; and Yang, M.-H. 2025.
\newblock Sa2VA: Marrying SAM2 with LLaVA for Dense Grounded Understanding of Images and Videos.
\newblock \emph{arXiv preprint arXiv:2501.04001}.

\bibitem[{Zhou et~al.(2024)Zhou, Gao, Zeng, and Lu}]{zhou2024dshmp}
Zhou, J.; Gao, J.; Zeng, H.; and Lu, H. 2024.
\newblock DsHmp: Densely Supervised Hierarchical Mask Propagation for Referring Video Object Segmentation.
\newblock \emph{arXiv preprint arXiv:2404.03645}.

\bibitem[{Zhu et~al.(2023)Zhu, Cheng, He, Li, Luo, Lu, Geng, and Xie}]{zhu2023tracking}
Zhu, J.; Cheng, Z.-Q.; He, J.-Y.; Li, C.; Luo, B.; Lu, H.; Geng, Y.; and Xie, X. 2023.
\newblock Tracking with human-intent reasoning.
\newblock \emph{arXiv preprint arXiv:2312.17448}.

\end{thebibliography}

\end{document}